\pdfoutput=1
\documentclass{article}

\PassOptionsToPackage{numbers, compress}{natbib}

\usepackage[preprint]{neurips_2026}

\usepackage[utf8]{inputenc} %
\usepackage[T1]{fontenc}    %
\usepackage{hyperref}       %
\usepackage{url}            %
\usepackage{booktabs}       %
\usepackage{amsfonts}       %
\usepackage{nicefrac}       %
\usepackage{microtype}      %
\usepackage{xcolor}         %

\usepackage{amsmath,amsfonts,bm}

\def\eqref#1{equation~\ref{#1}}

\def\1{\bm{1}}

\DeclareMathAlphabet{\mathsfit}{\encodingdefault}{\sfdefault}{m}{sl}
\SetMathAlphabet{\mathsfit}{bold}{\encodingdefault}{\sfdefault}{bx}{n}

\usepackage{xspace}

\usepackage{arydshln}
\usepackage{array, boldline, makecell, booktabs}
\usepackage[svgnames, table]{xcolor}
\usepackage{multirow}

\usepackage{pifont}

\usepackage{caption,subcaption,graphicx}

\usepackage{wrapfig}

\makeatletter
\NewDocumentCommand{\supptitle}{s}{
\twocolumn[{
  \begin{center}
    \vspace*{-0.3cm}
    \rule{\textwidth}{0.05cm}\\[0.1cm]
    \textbf{- Appendix -}\\[0.2cm]
    {\Large \textbf{\mytitle}}\\[0.1cm]
    \rule{\textwidth}{0.05cm}\\[0.3cm]
  \end{center}
}]
}
\makeatother

\newcommand{\eg}{\emph{e.g.,~}}
\newcommand{\ie}{\emph{i.e.,~}}

\newcommand{\cmark}{\textcolor{green}{\ding{51}}\xspace}%
\newcommand{\xmark}{\textcolor{red}{\ding{55}}\xspace}%

\definecolor{LightCyan}{rgb}{0.88,1,1}
\definecolor{Blue}{rgb}{0, 0.5, 1}
\definecolor{Green}{rgb}{0.0, 0.8, 0.0 }
\definecolor{Red}{rgb}{0.95, 0.55, 0.6}
\definecolor{Skyblue}{rgb}{0.6, 0.6, 0.95 }
\definecolor{Beige}{rgb}{0.96, 0.96, 0.86}
\definecolor{tablecolor}{RGB}{251, 198, 194}

\newif\ifshowcomments
\showcommentstrue

\newcommand{\alg}{{SignRot}\xspace}
\newcommand{\mytitle}{\alg: LLM Quantization with Massive Outlier-Aware Sign-Adjusted Rotation}

\usepackage[normalem]{ulem}
\useunder{\uline}{\ul}{}

\usepackage{tcolorbox}
\tcbuselibrary{skins, breakable}

\usepackage[hang,flushmargin]{footmisc}

\usepackage{enumitem}
\usepackage{cleveref}

\title{\mytitle}

\author{%
  Jaehun Gim \\
  Hanyang University \\
  \texttt{wogns3623@hanyang.ac.kr} \\
  \And
  Jaewoo Kim \\
  Hanyang University \\
  \texttt{shrimp0426@hanyang.ac.kr} \\
  \AND
  Gyuwan Kim \\
  University of California, Santa Barbara \\
  \texttt{gyuwankim@ucsb.edu} \\
  \And
  Seo Yeon Park\thanks{Corresponding author.} \\
  Hanyang University \\
  \texttt{seoyeonpark@hanyang.ac.kr} \\
}

\begin{document}

\maketitle

\begin{abstract}
While quantization compresses LLMs and accelerates inference, outlier activations impede efficient low-bit representation. In particular, massive outliers cause substantial performance degradation. Although rotation-based methods that rely on randomized Hadamard transforms for online inference aim to mitigate this, they can produce fragmented distributions with empty bins when a massive outlier dominates a token's activation. This fragmentation severely undermines uniform quantization performance. Moreover, massive outliers concentrate in the down-projection input, where the preceding nonlinearity requires the rotation to be applied at inference time. This constraint prevents training-based approaches such as SpinQuant from learning the online rotation at this position, as a learned dense matrix cannot use the Fast Walsh--Hadamard Transform and would require a full matrix multiplication during inference. To address this, we propose \alg (Sign-adjusted Rotation), which identifies massive outlier channels and constructs a sign adjustment vector that unifies the signs of the corresponding Hadamard row, collapsing the bimodal distribution into a unimodal profile amenable to quantization. Since the aligned outlier shifts the activation mean away from zero, we pair \alg with asymmetric activation quantization, whose zero-point absorbs this shift. 
As a lightweight, drop-in improvement compatible with any Hadamard-based quantization method, \alg requires only a single calibration forward pass and adds only element-wise sign flips at inference, without an additional matrix multiplication. 
Applying \alg to QuaRot achieves a WikiText2 perplexity of 5.88 and an average zero-shot reasoning accuracy of 72.73, surpassing SpinQuant on the LLaMA3-70B model.
Code is available at \url{https://github.com/HYU-ARK-Lab/SignRot}.
\end{abstract}

\section{Introduction}
While large language models (LLMs) have achieved remarkable performance across diverse domains \cite{thirunavukarasu2023large, cox2023conversational, roziere2023code}, their escalating scale presents substantial challenges in computational and memory requirements. Quantization has emerged as a promising efficiency technique due to its straightforward implementation and ability to maintain reliable performance under significant compression \cite{lin2024duquant, ashkboos2024quarot, liu2025spinquant}. In particular, post-training quantization (PTQ) \cite{frantar2023optq} has become a \emph{de facto} standard, as it requires only a minimal calibration dataset without extensive retraining, making it highly practical for large-scale deployment. PTQ has been applied to quantize weights \cite{frantar2023optq, chee2023quip}, activations \cite{pmlr-v202-xiao23c, ashkboos2024quarot}, and the KV cache \cite{hooper2024kvquant}. 
\begin{wrapfigure}{r}{0.5\linewidth}
    \centering
    \vspace*{15pt}
    \includegraphics[width=0.99\linewidth]{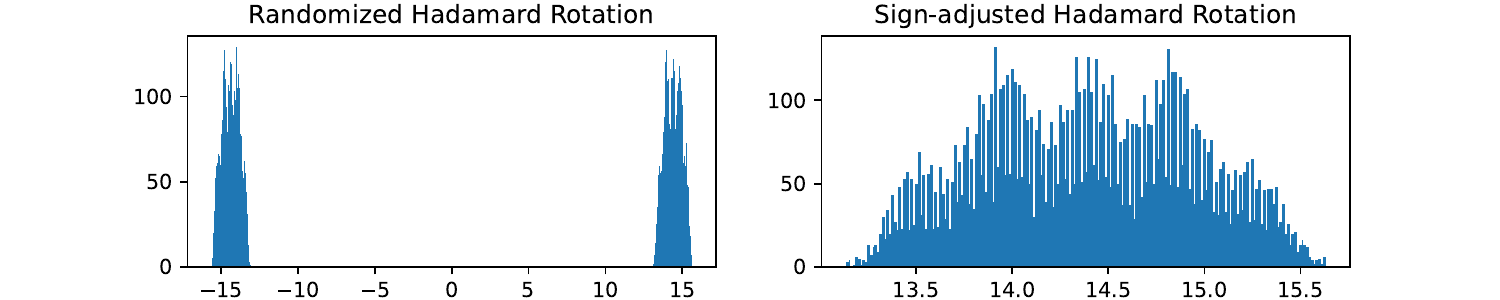}
    \caption{Distributions of token activations in LLaMA2-7B layer 2 down\_proj containing massive outliers after rotation using \emph{randomized} Hadamard and sign-adjusted Hadamard.}
    \label{fig:fragmented_distribution}
    \vskip -1em
\end{wrapfigure}

While PTQ methods are promising, the presence of large-magnitude outliers in activations, called massive outliers, remains a critical bottleneck, as they disproportionately distort the quantization process \cite{pmlr-v202-xiao23c, sun2024flatquant, ashkboos2024quarot, lin2024duquant}.  %
To mitigate this issue, rotation-based PTQ methods, which transform the activation space to suppress outliers before quantization, have been explored \cite{ashkboos2024quarot, liu2025spinquant, lin2024duquant}. These rotation matrices can generally be fused into adjacent weights offline, incurring no additional inference cost.
However, for layers following nonlinear activations, such as FFN down-projection layers, the rotation cannot be fused into adjacent weights and thus typically requires online application during inference. To mitigate the resulting computational overhead, prior work has explored the use of \emph {randomized} Hadamard matrices, which admit efficient implementation through the Fast Walsh--Hadamard Transform (FWHT) \cite{ashkboos2024quarot}.

However, we find that when a massive outlier dominates a token's input activation, applying \emph{randomized} Hadamard transforms produces a fragmented distribution with distinct positive and negative clusters, as shown in~\autoref{fig:fragmented_distribution}, which directly degrades quantization performance.
This is because the empty regions between these clusters inflate the activation quantization range and thus the step size.
This is particularly critical at the FFN down-projection, where massive outliers are known to concentrate~\cite{sun2024massive}, and where existing rotation-based PTQ methods are constrained to use randomized Hadamard transforms for efficiency.

To address this, we propose a simple yet effective activation-aware sign adjustment to the Hadamard transform, named \textbf{Sign}-adjusted \textbf{Rot}ation (\alg).
By applying column-wise sign flips that unify the signs of the Hadamard row corresponding to the identified massive outlier channel, the fragmented distribution is restored to a bell-shaped profile amenable to quantization. Because the aligned outlier shifts the rotated activation's mean away from zero, \alg pairs this adjustment with asymmetric activation quantization, whose zero-point absorbs the shift. 

\alg is a lightweight, drop-in improvement compatible with any Hadamard-based quantization method, including QuaRot \cite{ashkboos2024quarot}, SpinQuant \cite{liu2025spinquant}, and DartQuant \cite{shao2025dartquant}, and is applicable to both dense and MoE architectures. Since massive outlier channels are fixed across inputs \cite{sun2024massive}, the sign adjustment vector needs only to be computed once via a single calibration forward pass; at inference, \alg adds only an element-wise sign flip to the rotated activation, with no additional matrix multiplication.
Furthermore, \alg quantizes a 7B model in approximately 3 minutes with Round-to-Nearest (RTN), and around 15 minutes with GPTQ~\cite{frantar2023optq} on a single GPU, requiring no rotation training.

Our contribution is summarized as follows:
\begin{itemize}
\item We analytically characterize that randomized Hadamard transforms induce bimodal fragmentation in the presence of massive outliers, demonstrating that this fragmentation leads to quantization degradation at the FFN down-projection.
\item We propose \alg, a simple yet effective activation-aware sign adjustment to the Hadamard transform that directly addresses the fragmentation without requiring learned rotations. \alg is a lightweight, drop-in improvement compatible with any Hadamard-based quantization method, requiring only a single calibration forward pass.
\item \alg improves quantization performance across diverse settings, including both dense and MoE architectures. 
On LLaMA3-70B with GPTQ, \alg reduces the WikiText2 perplexity of QuaRot~\cite{ashkboos2024quarot} from 6.16 to 5.88 under the same asymmetric activation quantization, and applied on top of SpinQuant~\cite{liu2025spinquant}, which relies on learned rotations, it further reduces perplexity from 6.38 to 5.45 and improves average zero-shot accuracy from 71.15 to 73.52.
\end{itemize}

\begin{figure*}
\centering

\begin{subfigure}[t]{0.54\textwidth}
    \centering
    \caption{Self-attention}
    \includegraphics[width=\textwidth]{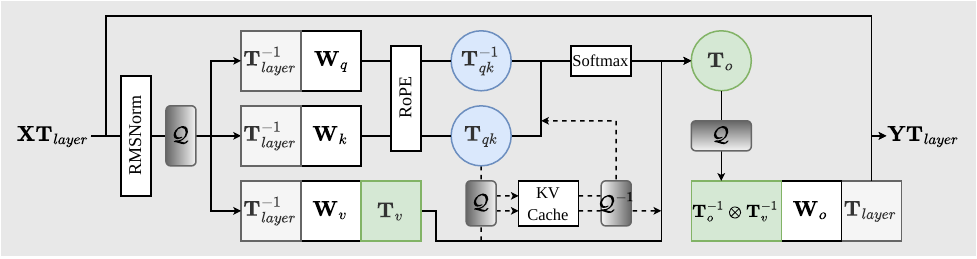}
    \label{fig:pipeline-self_attn}
\end{subfigure}\hfill
\begin{subfigure}[t]{0.44\linewidth}
    \centering
    \caption{Feed-forward network}
    \includegraphics[width=\textwidth]{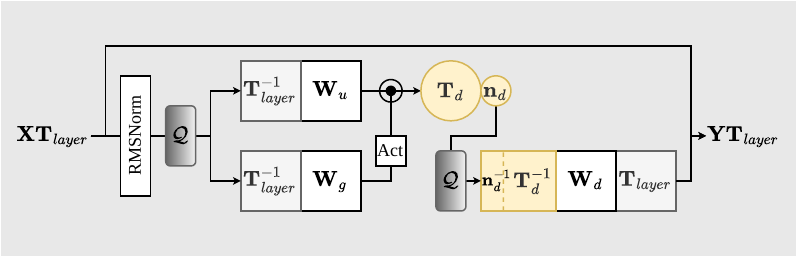}
    \vspace{-1.5ex}
    \label{fig:pipeline-mlp}
\end{subfigure}\hfill

\vspace{-1.3ex}
\begin{subfigure}[t]{\textwidth}
    \centering
    \includegraphics[width=\textwidth]{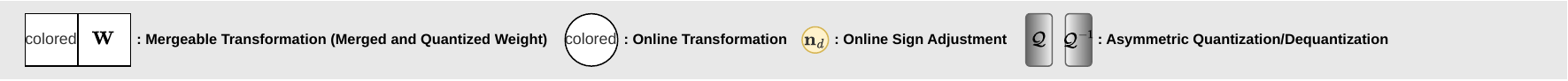}
    \label{fig:pipeline-notations}
\end{subfigure}
\vspace{-3ex}

\caption{\small The overview of our proposed method \alg in (a) self-attention and (b) feed-forward blocks, illustrated with a gated MLP (\eg LLaMA-style) as a representative architecture. Rotation matrices fall into two categories: mergeable transformation (\eg $\mathbf{T}_{layer}, \mathbf{T}_o \otimes \mathbf{T}_v$), which are fused into adjacent weights offline and incur no inference cost; and online transformation (\eg $\mathbf{T}_{d}$), which must be applied during inference due to preceding nonlinear operations. \alg introduces a sign adjustment vector $\mathbf{n}_d$ at $\mathbf{T}_d$, the down-projection input where massive outliers concentrate, merging $\operatorname{diag}(\mathbf{n}_d)^{-1}$ into $\mathbf{W}_d$ offline while applying $\mathbf{n}_d$ online to the intermediate activation $\mathbf{X}_d$. } 
\label{fig:pipeline}
\end{figure*}

\section{Preliminaries and Related Work}  %

\paragraph{Quantization formulations.}
Quantization is a technique that converts full-precision weights or activations (and optionally the KV cache) into low-precision (or discrete) values. This method is widely used to reduce the model size and inference latency. 
The general $b$-bit uniform quantization function $\mathcal{Q}_b$ maps an input matrix $\mathbf{X}$ into 
$\mathbf{X}_q = \mathcal{Q}_b(\mathbf{X}) = \Pi_{\Omega(b)}(\mathbf{X}/\Delta) + \mathbf{Z},$
where the subscript $_{q}$ mean quantized, $\Delta$ is the step-size, $\mathbf{Z}$ is the zero point for asymmetric quantization, $\Pi$ is a projection function onto the set of integers $\Omega(b) = \{0,1,\dots,2^b-1\}$ or $\Omega(b) = \{-2^{b-1}, \dots, 2^{b-1}-1\}$. 
For brevity, we omit $b$ and denote $\mathcal{Q}(\cdot)$ as the general quantization function in the rest of this paper.
The dequantized matrix $\widehat{\mathcal Q}(\mathbf X)=\mathbf{X}_{deq}$ is expressed as $\Delta \cdot (\mathbf{X}_q - \mathbf{Z})$, with scales and zero-points broadcast over their quantization groups.
During inference, a linear layer $\mathbf{Y} = \mathbf{XW}$ is approximated as $\mathbf{Y} \approx \mathbf{X}_{deq} \mathbf{W}_{deq}$.
In this work, we primarily use per-token asymmetric quantization for activations and per-channel symmetric quantization for weights.

\paragraph{Channel outlier and scaling methods}
Previous studies~\cite{dettmers2022gpt3,pmlr-v202-xiao23c,shao2023omniquant} have focused on handling channel outliers that appear in specific feature channels across entire token sequences.
These methods primarily leverage \emph{scale invariance}, which utilizes the fact that a linear layer $\mathbf{Y} = \mathbf{XW}$ remains mathematically equivalent when activations are scaled by a diagonal matrix $\mathbf{S}$ and weights are inversely scaled: $\mathbf{Y} = (\mathbf{XS})(\mathbf{S}^{-1}\mathbf{W})$.

While LLM.int8~\cite{dettmers2022gpt3} initially managed these outliers through mixed-precision decomposition, its implementation proved hardware-unfriendly.
To address this, SmoothQuant~\cite{pmlr-v202-xiao23c} introduced a technique to migrate the quantization difficulty of channel outliers from activations to weights via channel-wise scaling. 
This direction was further refined by OmniQuant~\cite{shao2023omniquant}, which improved quantization performance by learning optimal channel-wise scaling factors.

However, scaling methods that only adjust the per-channel magnitude cannot effectively handle a newly discovered type of outlier~\cite{sun2024massive,liu2024intactkv} called the massive outlier.
These outliers concentrate in specific tokens and channels, particularly in FFN layers, and are known to be significant in quantizing activations to 4 bits or less~\cite{lin2024duquant}.

\paragraph{Massive outlier and rotation methods}
To mitigate the effects of massive outliers, recent studies~\cite{ashkboos2024quarot,liu2025spinquant,shao2025dartquant} leverage \emph{transformation invariance}.
Unlike scaling, which only performs element-wise adjustment, rotation-based methods use an orthogonal transformation $\mathbf{R}$ to redistribute the outlier energy across all available channels: $\mathbf{Y} = (\mathbf{XR})(\mathbf{R}^{-1}\mathbf{W})$.
The applicability and optimization of these rotations are fundamentally governed by their position within the Transformer~\cite{vaswani2017attention} block.

When a rotation follows a linear operation, it can be fused directly into adjacent weights during the PTQ phase, incurring no additional inference cost.
This offline mergeability allows for intensive optimization through training to maximize accuracy.
For instance, SpinQuant~\cite{liu2025spinquant} optimizes the rotation matrix using CayleySGD~\cite{li2020efficient}. 
DartQuant~\cite{shao2025dartquant} further improves the efficiency of rotation matrix optimization by constraining the post-rotation activation distribution, which reduces both computational cost and overfitting risk.

In contrast, rotations that follow nonlinear operations, such as the FFN down-projection input (\ie $\text{T}_d$ in~\autoref{fig:pipeline}, where a LLaMA-style architecture is used as a representative example), cannot be fused and must be applied during inference. 
To maintain inference efficiency at these positions, existing methods are constrained to use the Randomized Hadamard Transform (RHT) to leverage the Fast Walsh--Hadamard Transform (FWHT)~\cite{fino1976unified}. 
Various methodologies have been employed to implement this framework. QuaRot~\cite{ashkboos2024quarot} serves as a foundation for training-free rotation, effectively removing outliers using RHT, while DFRot~\cite{xiang2024dfrot} refines the rotation using a small calibration dataset to minimize quantization errors in a near-closed-form manner.

While these training-free methods offer fast setup using a minimal calibration step, they remain agnostic to the specific polarity of massive outliers. 
As we will analyze in Section~\ref{sec:fragmentation}, this disregard for outlier polarity induces a systematic bimodal fragmentation when using RHT at the FFN down-projection. This serves as a critical source of quantization degradation that cannot be resolved without considering the signs of the outliers.

\section{Analysis of Bimodal Fragmentation in RHT under Massive Outliers}
\label{sec:fragmentation}

The \emph{randomized} Hadamard transform (RHT) combines independent random signs with the uniform structure of the Hadamard matrix. In the absence of dominant outliers, its output can have a bell-shaped distribution favorable for quantization~\cite{tseng2024quip}. The Gaussian approximation used here requires the conditions discussed below.
The Gaussian approximation under RHT relies on the Central Limit Theorem (CLT). Independent random signs yield zero-mean summands. In this setting, the Lindeberg condition requires that no single summand contribute a non-negligible fraction of the total variance in the limit~\cite{berckmoes2018approximate}.
This assumption can fail when one channel retains a non-vanishing fraction of the activation energy; that channel then contributes an independent two-valued term, so the limit is not Gaussian.

Formally, let $\mathbf{x}_t \in \mathbb{R}^{C_{in}}$ be the activation vector for token $t$, with signed outlier value $M_t=x_{t,c_{max}}$ at channel $c_{max}$.
When the RHT is applied, the output is computed as $\mathbf{y}_t = \mathbf{x}_t \cdot \text{diag}(\mathbf{r}) \cdot \mathbf{H}$, where $\mathbf{H}$ is an orthonormalized Hadamard matrix and $r_i \in \{\pm 1\}$ are independent random signs.
The $j$-th element of the transformed vector decomposes as follows: 
\begin{equation}
\label{eq:decomposition}
    y_{t,j} = M_t \cdot r_{c_{max}} h_{c_{max},j} + \sum_{i \neq c_{max}} x_{t,i} r_i h_{i,j}
\end{equation}
where $h_{i,j}$ is the $(i,j)$-th entry of the Hadamard matrix $\mathbf{H}$ and $x_{t,i}$ is the $i$-th element of $\mathbf{x}_t$.
For a fixed output channel $j$, the residual has variance $\sigma_{rest}^2 = C_{in}^{-1}\sum_{i \neq c_{max}} x_{t,i}^2$ over the random signs. When the residual satisfies its own Lindeberg condition, we approximate it by $\mathcal{N}(0,\sigma_{rest}^2)$. The independent outlier term takes values $\pm M_t/\sqrt{C_{in}}$ with equal probability, giving the Gaussian-mixture approximation
\begin{equation}
\label{eq:bimodal_mixture}
    y_{t,j} \;\overset{\mathrm{approx.}}{\sim}\; \frac{1}{2}\mathcal{N}\!\left(+\frac{M_t}{\sqrt{C_{in}}}, \sigma_{rest}^2\right) + \frac{1}{2}\mathcal{N}\!\left(-\frac{M_t}{\sqrt{C_{in}}}, \sigma_{rest}^2\right).
\end{equation}

This mixture describes a distribution over random signs, not an exact law for the channels of one fixed transform. For a balanced outlier row, half the channels receive each shift; interpreting their histogram as this mixture additionally assumes approximately Gaussian residuals within both groups (\autoref{sec:bimodality_threshold}).
For the full sum, the Lindeberg condition reduces to $\max_i x_{t,i}^2 / \sum_i x_{t,i}^2 \to 0$. Its violation is asymptotic; the finite threshold below is exact only within the Gaussian-mixture model.
An equally weighted mixture of two Gaussians with a common variance $\sigma^2$ is bimodal if and only if their means are more than $2\sigma$ apart~\cite{schilling2002bimodal}.
Applying this to~\autoref{eq:bimodal_mixture}, \ie $|M_t|/\sqrt{C_{in}} > \sigma_{rest}$, yields
\begin{equation}
\label{eq:bimodality_threshold}
    M_t^2 > \sum_{i \neq c_{max}} x_{t,i}^2,
\end{equation}
that is, the outlier channel must carry more energy than all remaining channels combined.
Since this requires $M_t^2 > \|\mathbf{x}_t\|_2^2/2$, at most one channel per token can satisfy~\autoref{eq:bimodality_threshold}. 
We refer to such a channel as the dominant massive outlier channel of the token; it is the channel that \alg targets. Accordingly, a massive outlier token is defined as a token whose largest channel satisfies~\autoref{eq:bimodality_threshold} and reaches at least half of the peak activation magnitude of the layer, provided that this peak is at least 100. This energy-based criterion differs from the magnitude-based criterion of Sun et al.~\cite{sun2024massive}; \autoref{sec:ma_statistics} compares the two criteria.

For the massive outlier token of LLaMA2-7B in~\autoref{tab:activation_distribution_statistics} ($C_{in}=11{,}008$), the threshold is $\sqrt{C_{in}}\,\sigma_{rest} \approx 58$, whereas the observed outlier is $M_t = 1510$, about $26\times$ larger, consistent with the clearly separated clusters in~\autoref{fig:fragmented_distribution}. 
The full derivation is provided in~\autoref{sec:bimodality_threshold}. Since the threshold is derived under the two-Gaussian approximation, energy dominance should be read as an indicator of fragmentation rather than an exact test for empirical bimodality; in particular, several large residual channels can invalidate the two-component model. 
As illustrated in~\autoref{fig:fragmented_distribution}, bimodal fragmentation can degrade quantization performance by widening the activation range.
For an input activation token $\mathbf{x}_t$ quantized with a shared step size $\Delta_t$, nearest rounding without clipping gives $|\widehat{\mathcal Q}(x_{t,j})-x_{t,j}|\leq\Delta_t/2$ for each channel. The mean squared error (MSE) of the quantized input is therefore
\begin{equation}
\label{eq:token_quantization_mse}
    \operatorname{MSE}_t
    := \frac{1}{C_{in}}\left\|\widehat{\mathcal Q}(\mathbf{x}_t)-\mathbf{x}_t\right\|_2^2
    = \frac{1}{C_{in}}\sum_{j=1}^{C_{in}}\left(\widehat{\mathcal Q}(x_{t,j})-x_{t,j}\right)^2
    \leq \frac{\Delta_t^2}{4}.
\end{equation}
Since the two component means are separated by $2|M_t|/\sqrt{C_{in}}$, the per-token range grows by at least this amount in addition to the spread of the residual, which increases $\Delta_t$ at a fixed bit width and hence the rounding error in~\autoref{eq:token_quantization_mse}. 
The gap between the clusters additionally leaves the central quantization bins unused, while the realized MSE also depends on grid alignment and clipping. 
Although massive outlier tokens are rare, the fragmentation they cause substantially increases the quantization error of the layer in which they first appear (\autoref{sec:ma_statistics}).
The range inflation itself, however, follows directly from the sign pattern of the outlier row, and removing it is the goal of the sign adjustment introduced next. 
We provide activation distribution statistics for massive outlier tokens and non-outlier tokens in~\autoref{sec:activation_distribution_statistics}.

\section{Proposed Method: \alg}
\label{sec:methodology}
\begin{wrapfigure}[13]{r}{0.5\linewidth}
    \centering
    \vskip -3em
    \includegraphics[width=0.99\linewidth]{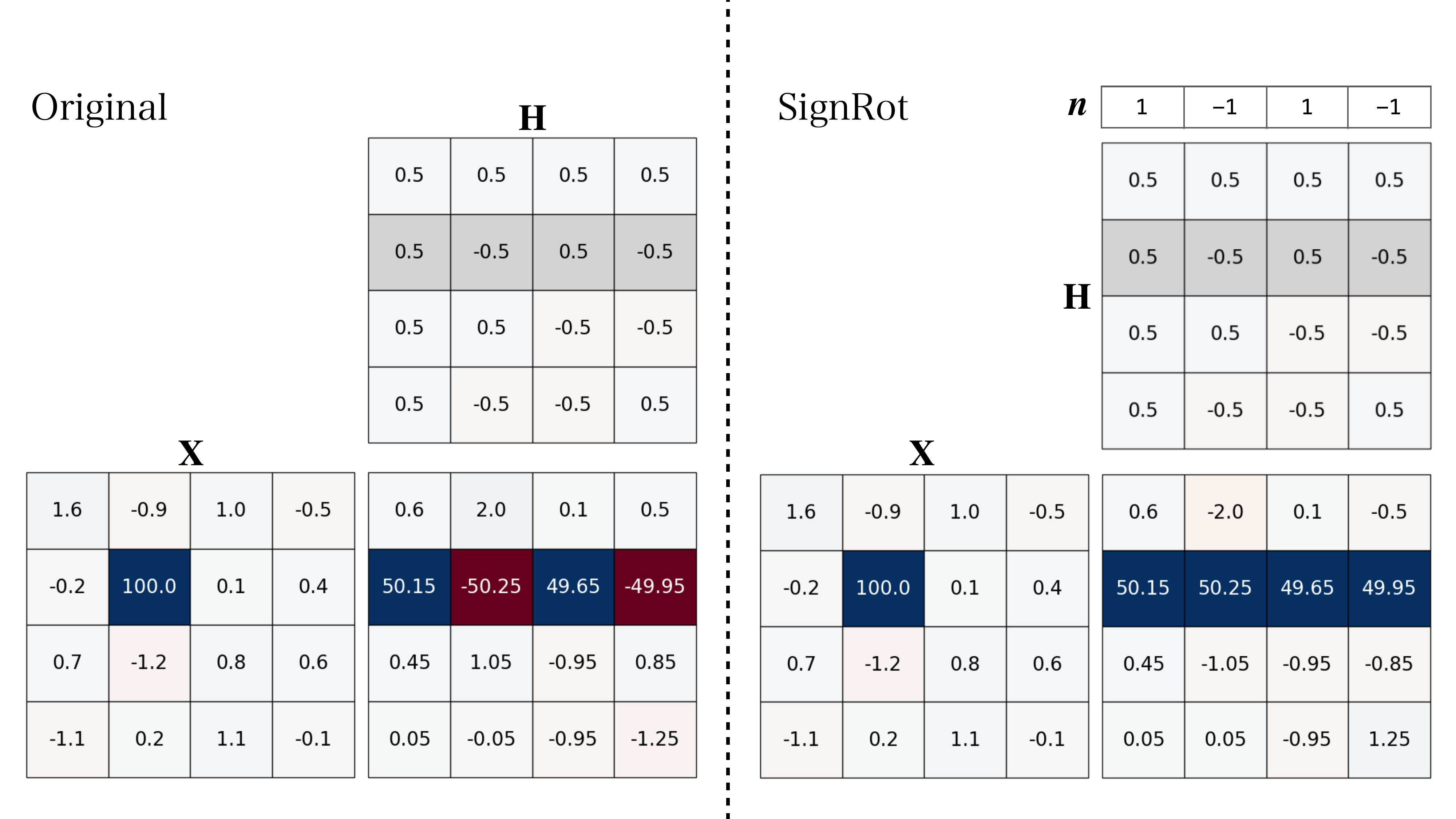}
    \caption{\small Effect of sign adjustment on the {\em randomized} Hadamard matrix under a massive outlier. Without adjustment (left), large outlier activation produces fragmented outputs of mixed sign. With sign alignment via $\operatorname{diag}(\boldsymbol{n})$ (right), the outlier contribution becomes a uniform positive shift, collapsing the bimodal distribution.} %
    \label{fig:sign_adjusted_hadamard}
    \vskip -1em
\end{wrapfigure}
The bimodal fragmentation analyzed above arises from the alternating signs of the Hadamard entries $h_{c_{max},j} \in \{\pm 1/\sqrt{C_{in}}\}$ corresponding to the outlier channel $c_{max}$. We therefore propose \textbf{Sign}-adjusted \textbf{Rot}ation (\alg), which unifies the signs of the $c_{max}$-th row of $\mathbf{H}$, such that the selected outlier contribution becomes a constant shift $M_t/\sqrt{C_{in}}$ across all output channels. If the adjusted residual is approximately Gaussian, the resulting distribution is approximately unimodal.

\subsection{Identifying the massive outlier channel}
To identify the target channel, we examine the activation matrix $\mathbf{X} \in \mathbb{R}^{T \times C_{in}}$ over a calibration set and locate the channel with the largest absolute activation:
\begin{equation}
    c_{\max} = \arg\max_{c} \left( \max_{t} |x_{t,c}| \right).
\end{equation}
We identify one channel in a calibration forward pass and reuse it for subsequent inputs.
This choice follows the observed stability of massive outlier channels~\cite{sun2024massive}: in the first massive outlier layer of every model we examine, all massive outlier tokens share the same dominant massive outlier channel, which holds 87--99.9\% of the token energy (\autoref{tab:early_ma_layers}, \autoref{sec:ma_statistics}). When a token carries further massive outlier channels, \ie channels that each carry more energy than all smaller channels combined (\autoref{eq:bimodality_threshold} applied recursively), \alg still removes the split induced by the selected channel (\autoref{sec:multi_ma}). Aligning this single channel therefore removes most of the fragmentation, although some later layers do not share the same dominant massive outlier channel (\autoref{sec:limitations}).

\subsection{Sign-adjusted Hadamard transformation}
\label{sec:sign_adjusted_hadamard}
Given $c_{max}$, we define the sign adjustment vector $\mathbf{n} \in \{+1,-1\}^{C_{in}}$ as follows:
\begin{equation}
\label{eq:sign_adjust}
    {n}_{j} := \text{sign}(h_{c_{max},j})= \begin{cases}
        1 & \text{if } h_{c_{\max}, j} \ge 0, \\
        -1 & \text{otherwise},
    \end{cases}
\end{equation}
We then define the sign-adjusted Hadamard transformation matrix $\mathbf{H}_{\mathrm{adj}} \in \mathbb{R}^{C_{in} \times C_{in}}$ as follows:
\begin{equation}
\label{eq:full_transform_matrix}
    \mathbf{H}_{\mathrm{adj}} = \mathbf{H}\, \operatorname{diag}(\mathbf{n}),
\end{equation}
where right-multiplication by $\operatorname{diag}(\mathbf{n})$ flips the sign of each column $j$ by $n_j$, so that every entry of the $c_{\max}$-th row of $\mathbf{H}_{\mathrm{adj}}$ becomes $+1/\sqrt{C_{in}}$.
We demonstrate this effect in~\autoref{fig:sign_adjusted_hadamard}. Without sign adjustment, the outlier row (\ie containing the massive activation $\approx 100.0$) produces mixed-sign outputs that directly reflect the alternating signs of the corresponding Hadamard row (\autoref{fig:sign_adjusted_hadamard} left). After applying $\operatorname{diag}(\mathbf{n})$, the same row produces uniformly positive outputs (\autoref{fig:sign_adjusted_hadamard} right). 
Consequently, the selected outlier term $M_t h_{c_{max},j} n_j = M_t/\sqrt{C_{in}}$ becomes a constant shift for all $j$, removing this channel's sign-induced split.
In practice, we apply this adjustment at the down-projection input ($\mathbf{T}_d$ in~\autoref{fig:pipeline}), where massive outliers concentrate~\cite{sun2024massive}. We denote the corresponding vector as $\mathbf{n}_d$.

\paragraph{Quantization with sign-adjusted Hadamard transformation.} Since $\operatorname{diag}(\mathbf{n})$ is an involutory matrix (\ie $\operatorname{diag}(\mathbf{n})^{-1}= \operatorname{diag}(\mathbf{n})$), the inverse is given by $\mathbf{H}_{\mathrm{adj}}^{-1} = \operatorname{diag}(\mathbf{n})\,\mathbf{H}^{-1}$. 
This allows the adjusted transformation to be inserted into a linear layer without altering the computation since $\mathbf{Y}=\mathbf{X}\mathbf{W} = (\mathbf{X}\mathbf{H}_{\mathrm{adj}})(\mathbf{H}_{\mathrm{adj}}^{-1}\mathbf{W})$ holds. For clarity, the random-sign factor $\operatorname{diag}(\mathbf r)$ is omitted here. Quantization is applied to each factor independently; $\mathbf Y_q$ denotes the resulting integer accumulation:
\begin{equation}
\label{eq:quant_layer}
    \mathbf{Y}_{q} = \mathcal{Q}\!\left(\mathbf{X}\mathbf{H}_{\mathrm{adj}}\right) \mathcal{Q}\!\left(\mathbf{H}_{\mathrm{adj}}^{-1}\mathbf{W}\right).
\end{equation}

\paragraph{Asymmetric--Symmetric matrix multiplication}
\alg pairs sign adjustment with asymmetric activation quantization to represent the resulting shifted distribution. Sign adjustment changes the signs of the Hadamard outputs while preserving their absolute values; it therefore leaves the max-absolute scale of symmetric quantization unchanged. Asymmetric quantization uses a zero-point to accommodate the shift and can allocate quantization levels to the shifted range. 
Asymmetric activation quantization is therefore an integral part of \alg rather than an optional refinement. 
To maintain the efficiency of activation quantization, we use Asymmetric--Symmetric matrix multiplication (Asym--Symm matmul, where activations are quantized asymmetrically and weights symmetrically) defined as follows:
\begin{equation}
\label{eq:asym_symm_matmul}
    \mathbf{Y}_{deq} = \Delta_{\mathbf{X}} (\mathbf{X}_q \mathbf{W}_q - \mathbf{Z_X}\mathbf{W}_q) \Delta_{\mathbf{W}}
\end{equation}
where $\mathbf{Y}_{deq}$ denotes the dequantized output, $\Delta_{\mathbf{X}}$ and $\Delta_{\mathbf{W}}$ are diagonal matrices of per-token activation scales and per-output-channel weight scales, respectively, and $\mathbf{Z_X}$ contains the activation zero-points.
The first term in~\autoref{eq:asym_symm_matmul} is an integer dot product identical to Symmetric--Symmetric matrix multiplication (Symm--Symm matmul, where both activations and weights are quantized symmetrically).
For per-token activation quantization, $\mathbf{Z_X}=\mathbf{z_X}\mathbf{1}^{\top}$, where $\mathbf{z_X}$ is a column vector of token zero-points and $\mathbf{1}\in\mathbb R^{C_{in}}$ is the all-ones vector. The second term therefore factorizes as $\mathbf{z_X}(\mathbf{1}^{\top}\mathbf{W}_q)$. The weight column sum $\mathbf{1}^{\top}\mathbf{W}_q$, with $\mathbf{W}_q\in\mathbb Z^{C_{in}\times C_{out}}$, is input-independent and precomputed offline. During inference, Asym--Symm matmul computes the rank-1 outer product with $\mathbf{z_X}$ and subtracts it from the integer matrix product. This correction costs $O(T\,C_{out})$, compared with $O(T\,C_{in}C_{out})$ for the matrix multiplication. Asymmetric quantization also estimates a per-token zero-point alongside the scale required by symmetric quantization.

\section{Experiments}

\subsection{Experimental setup}
\paragraph{Models and datasets}
We apply \alg on the LLaMA2 (7B, 13B, 70B)~\cite{touvron2023llama2} and LLaMA3 (8B, 70B)~\cite{dubey2024llama} model families, and additionally on Mistral-7B-v0.3~\cite{jiang2023mistral7b} and Qwen3-30B-A3B~\cite{yang2025qwen3technicalreport}. To assess our method, we evaluate quantized models in language modeling and reasoning tasks.
We report perplexity on WikiText2~\cite{merity2016wiki} and C4~\cite{raffel2020c4} datasets to evaluate language modeling. 
Furthermore, we measure the accuracy on six zero-shot reasoning tasks: ARC (Easy and Challenge)~\cite{clark2018arc}, HellaSwag~\cite{zellers2019hellaswag}, LAMBADA(OpenAI)~\cite{radford2019language}, PIQA~\cite{bisk2020piqa}, and WinoGrande~\cite{sakaguchi2021winogrande}.

\paragraph{Baselines}
We evaluate \alg{} by applying it on top of three rotation-based methods that employ online Hadamard transforms: QuaRot~\cite{ashkboos2024quarot}, SpinQuant~\cite{liu2025spinquant}, and DartQuant~\cite{shao2025dartquant}. QuaRot applies randomized Hadamard transforms without any training, while SpinQuant and DartQuant further optimize the rotation matrices to minimize quantization error. 

\paragraph{Implementation details}
\label{sec:implementation_details}
We implemented \alg using HuggingFace Transformers~\cite{wolf2019huggingface} and PyTorch~\cite{paszke2019pytorch}.
On a single A100 GPU, \alg only adds 5 minutes of quantization time for the LLaMA2-70B model, which is negligible compared to the total quantization time. The peak memory usage during quantize LLaMA2-70B is max 7 GB for the rotation process and max 40 GB for the GPTQ stage. We use CUTLASS~\cite{nvidia_cutlass} for 4-bit matrix multiplication and FlashInfer~\cite{flashinfer} for KV cache quantization to evaluate the inference performance of our fully quantized model. Total quantization time is measured on a single NVIDIA A100 GPU, while inference speedup and memory usage are measured on an NVIDIA RTX 3090 GPU.

\paragraph{Quantization configurations}
\label{sec:quantization_settings}
We report results using both Round-to-Nearest (RTN) and GPTQ~\cite{frantar2023optq} for weight quantization; our GPTQ implementation uses 128 calibration samples randomly selected from the WikiText-2 dataset. To address the memory-intensive nature of self-attention, we apply group-wise asymmetric quantization to the KV cache with a group size of the head dimension, following~\cite{ashkboos2024quarot}. For each baseline, +\alg uses the same quantization configuration as the baseline, except that +\alg applies per-token asymmetric activation quantization and its clipping ratios are selected by hyperparameter tuning. QuaRot uses per-token symmetric activation quantization following its official setting~\cite{ashkboos2024quarot}, whereas SpinQuant and DartQuant already use asymmetric activation quantization. \autoref{sec:detailed_quant_config} gives the detailed quantization configuration of each baseline and of \alg.
Sign adjustment vectors are computed using the first sample of the GPTQ calibration dataset during the GPTQ calibration phase. For RTN, we compute the sign adjustment vector using a single sample randomly extracted from the WikiText-2 dataset. We confirm that the choice of calibration data has minimal impact on performance in \autoref{tab:calibration_data_insensitivity}.

\subsection{Main results}

\begin{table*}[!t]
  \newcommand{\hlt}{\rowcolor{tablecolor!50}}
  \centering
  \caption{
    Comparison of the perplexity score on WikiText2 and C4 (\textbf{PPL}), and averaged accuracy on six zero-shot reasoning tasks (\textbf{Avg. 0-shot}) for LLaMA families. `+\alg' indicates that our method has been applied. Full results are in the \autoref{sec:full_results}. Note that QuaRot uses symmetric activation quantization. \protect
  }
  \resizebox{\linewidth}{!}{
  \begin{tabular}{l c ccc ccc ccc ccc ccc}
    \toprule
    \multirow{3}{*}{\textbf{Method}} & \multirow{3}{*}{\textbf{W Quantizer}} & \multicolumn{3}{c}{\textbf{2-7B}} & \multicolumn{3}{c}{\textbf{2-13B}} & \multicolumn{3}{c}{\textbf{2-70B}} & \multicolumn{3}{c}{\textbf{3-8B}} & \multicolumn{3}{c}{\textbf{3-70B}} \\
    \cmidrule(lr){3-5}\cmidrule(lr){6-8}\cmidrule(lr){9-11}\cmidrule(lr){12-14}\cmidrule(lr){15-17}
    & & \textbf{Wiki2}&\textbf{C4}&\textbf{Avg.} & \textbf{Wiki2}&\textbf{C4}&\textbf{Avg.} & \textbf{Wiki2}&\textbf{C4}&\textbf{Avg.} & \textbf{Wiki2}&\textbf{C4}&\textbf{Avg.} & \textbf{Wiki2}&\textbf{C4}&\textbf{Avg.} \\
    & & \textbf{PPL$\downarrow$} & \textbf{PPL$\downarrow$} & \textbf{0-shot$\uparrow$}
    & \textbf{PPL$\downarrow$} & \textbf{PPL$\downarrow$} & \textbf{0-shot$\uparrow$}
    & \textbf{PPL$\downarrow$} & \textbf{PPL$\downarrow$} & \textbf{0-shot$\uparrow$}
    & \textbf{PPL$\downarrow$} & \textbf{PPL$\downarrow$} & \textbf{0-shot$\uparrow$}
    & \textbf{PPL$\downarrow$} & \textbf{PPL$\downarrow$} & \textbf{0-shot$\uparrow$} \\
    \midrule
    FP16        & -    &  5.47 & 7.26 & 69.79  &  4.88 & 6.73 & 72.55  &  3.32 & 5.71 & 77.05  &  6.14 & 9.45 & 73.23  &  2.86 & 7.17 & 79.95 \\
    \midrule
    QuaRot      & RTN  &  8.56 & 11.86 & 57.73  &  6.10 & 8.67 & 66.25  &  4.14 & 6.42 & 73.47  &  10.60 & 17.19 & 61.34  &  55.44 & 79.48 & 35.36 \\
    \hlt QuaRot+\alg & RTN  &  7.93 & 11.21 & 60.43  &  6.00 & 8.48 & 67.51  &  4.05 & 6.32 & 73.86  &  9.68 & 15.72 & 64.17  &  41.47 & 58.26 & 38.52 \\ \cdashline{1-17}
    QuaRot      & GPTQ &  6.10 & 8.32 & 65.01  &  5.40 & 7.54 & 68.91  &  3.79 & 6.12 & 75.68  &  8.16 & 13.38 & 65.79  &  6.60 & 12.87 & 70.45 \\
    \hlt QuaRot+\alg & GPTQ &  6.09 & 8.31 & 67.19  &  5.32 & 7.43 & 70.70  &  3.74 & 6.04 & 75.68  &  7.85 & 12.74 & 67.27  &  5.88 & 11.24 & 72.73 \\
    \midrule
    SpinQuant      & GPTQ &  6.08 & 8.25 & 66.27  &  5.39 & 7.46 & 69.94  &  3.73 & 6.04 & 75.65  &  7.85 & 12.71 & 67.96  &  6.38 & 13.98 & 71.15 \\
    \hlt SpinQuant+\alg & GPTQ &  6.09 & 8.27 & 67.07  &  5.37 & 7.45 & 70.84  &  3.72 & 6.04 & 75.79  &  7.87 & 12.60 & 69.36  &  5.45 & 10.54 & 73.52 \\
    \midrule
    DartQuant      & GPTQ &  5.97 & 8.07 & 66.98  &  5.32 & 7.39 & 70.41  &  3.66 & 6.01 & 75.61  &  7.49 & 12.16 & 69.27  &  4.93 & 9.45 & 76.39 \\
    \hlt DartQuant+\alg & GPTQ &  5.94 & 8.04 & 67.27  &  5.31 & 7.37 & 70.59  &  3.66 & 6.00 & 75.86  &  7.40 & 12.05 & 69.03  &  4.86 & 9.26 & 76.02 \\
    \bottomrule
  \end{tabular}}
  \vspace{1ex}
  \label{tab:main_acc}
\end{table*}

\paragraph{Results on LLaMA families.}
We report the main results in~\autoref{tab:main_acc}, where \alg is applied on top of three rotation-based baselines, respectively, on LLaMA2 (7B/13B/70B) and LLaMA3 (8B/70B) under both RTN and GPTQ weight quantization. We evaluate language modeling perplexity on WikiText2 and C4, and zero-shot reasoning task accuracy. 

For language modeling tasks, QuaRot+\alg generally reduces perplexity over the QuaRot baseline. The improvement is especially pronounced for LLaMA3-70B, where Wiki2 perplexity drops from 6.60 to 5.88 under GPTQ (from 6.16 to 5.88 under the same asymmetric activation quantization; \autoref{tab:ablation_symmetric_70b}), surpassing SpinQuant~\cite{liu2025spinquant} (6.38). For LLaMA2-13B with GPTQ, QuaRot+\alg achieves a Wiki2 PPL of 5.32, matching DartQuant~\cite{shao2025dartquant} and outperforming SpinQuant (5.39). On LLaMA3-8B, QuaRot+\alg+GPTQ reduces PPL from 8.16 to 7.85, matching SpinQuant and narrowing the gap to DartQuant (7.49). Applying \alg on top of SpinQuant and DartQuant also yields perplexity reductions in most settings, demonstrating that sign adjustment is broadly complementary to other rotation-based methods.

For zero-shot reasoning tasks, we make the following observations: under RTN quantization, QuaRot+\alg generally outperforms QuaRot, \ie achieving an average accuracy improvement of up to 2.83 points on LLaMA3-8B. On LLaMA2, this gain decreases with model size (2.70, 1.26, and 0.39 points for 7B, 13B, and 70B), consistent with the greater robustness of larger models to quantization~\cite{li2020train, lee2025exploring}: the accuracy drop from FP16 to QuaRot under RTN decreases from 12.06 to 6.30 and 3.58 points, leaving less quantization error for \alg to recover. Under GPTQ, QuaRot+\alg surpasses SpinQuant~\cite{liu2025spinquant} by 0.92 points on LLaMA2-7B (67.19 vs.\ 66.27) and by 0.76 points on LLaMA2-13B (70.70 vs.\ 69.94), even though SpinQuant relies on learned rotation matrices. 
On the larger LLaMA3-70B model, QuaRot+\alg{}+GPTQ improves average accuracy from 70.45 to 72.73 (from 72.28 to 72.73 under the same asymmetric activation quantization; \autoref{tab:ablation_symmetric_70b}), again exceeding SpinQuant (71.15). Furthermore, applying \alg{} to SpinQuant and DartQuant~\cite{shao2025dartquant} baselines yields additional gains in most settings, demonstrating the broad applicability of our approach.

\begin{table*}[t]
  \newcommand{\hlt}{\rowcolor{tablecolor!50}}
  \centering
  \caption{
    Zero-shot reasoning task accuracy (\textbf{0-shot}) and perplexity scores on WikiText2 and C4 (\textbf{PPL}) for 4-bit weight, activation, and KV-cache quantized Mistral and Qwen3. Note that QuaRot uses symmetric activation quantization. \protect }
  \label{tab:ablation_mistral}
  \resizebox{0.95\linewidth}{!}{
  \begin{tabular}{c l c cc ccccccc}
    \toprule
    \multirow{2}{*}{\textbf{Model}} 
& \multirow{2}{*}{\textbf{Method}} 
& \multirow{2}{*}{\textbf{W Quantizer}} 
& \textbf{Wiki2} 
& \textbf{C4} 
& \textbf{ARC-C} 
& \textbf{ARC-E}
& \textbf{HellaSwag}
& \textbf{LAMBADA}
& \textbf{PIQA}
& \textbf{Winogrande}
& \textbf{Avg.} \\
    
& & 
& \textbf{PPL$\downarrow$}
& \textbf{PPL$\downarrow$}
& \textbf{0-shot$\uparrow$}
& \textbf{0-shot$\uparrow$}
& \textbf{0-shot$\uparrow$}
& \textbf{0-shot$\uparrow$}
& \textbf{0-shot$\uparrow$}
& \textbf{0-shot$\uparrow$}
& \textbf{0-shot$\uparrow$} \\
    \midrule
    \multirow{3}{*}{\textbf{Mistral-7B-v0.3}}
    & FP16        & -   & 5.32 & 8.47 & 52.30 & 78.20 & 80.41 & 75.32 & 82.26 & 73.88 & 73.73 \\
    \cmidrule{2-12}
    & QuaRot      & GPTQ & 5.80 & 9.19 & 48.89 & 76.13 & 78.09 & 73.18 & 79.86 & 72.37 & 71.42 \\
    & QuaRot+\alg & GPTQ & 5.71 & 9.09 & 48.72 & 76.09 & 78.95 & 72.64 & 80.30 & 72.45 & 71.53 \\ 
    \midrule
    
    \multirow{5}{*}{\textbf{Qwen3-30B-A3B}}
    & FP16        & -    & 8.70  & 14.05 & 56.57 & 79.25 & 77.63 & 64.76 & 80.47 & 69.38 & 71.34 \\ 
    \cmidrule{2-12}
    & QuaRot      & RTN  & 11.70 & 18.21 & 47.18 & 68.90 & 69.00 & 52.73 & 75.57 & 63.38 & 62.79 \\
    & QuaRot+\alg & RTN  & 11.21 & 17.63 & 48.04 & 69.70 & 70.04 & 53.97 & 75.30 & 62.04 & 63.18 \\
    \cmidrule{2-12}
    & QuaRot      & GPTQ & 10.16 & 16.22 & 48.89 & 73.82 & 71.51 & 57.25 & 75.68 & 63.54 & 65.11 \\
    & QuaRot+\alg & GPTQ & 9.79  & 15.93 & 47.70 & 72.10 & 72.33 & 60.99 & 76.99 & 64.72 & 65.80 \\
    \bottomrule
  \end{tabular}}
\end{table*}

\paragraph{Results on other models}
To validate the generalizability of our method, we apply \alg to two additional models: Mistral-7B-v0.3~\cite{jiang2023mistral7b} and Qwen3-30B-A3B~\cite{yang2025qwen3technicalreport}, and show results in~\autoref{tab:ablation_mistral}. \alg generally improves aggregate performance over the QuaRot baseline: it lowers perplexity in all settings and raises the average zero-shot accuracy, although some individual tasks decrease (\eg LAMBADA on Mistral-7B-v0.3 and ARC on Qwen3-30B-A3B with GPTQ). On Mistral-7B-v0.3, \alg with GPTQ reduces the WikiText2 perplexity from 5.80 to 5.71 while keeping a similar average accuracy (71.53 vs. 71.42). On Qwen3-30B-A3B, \alg with GPTQ outperforms QuaRot with GPTQ by 0.69 points in average accuracy, suggesting that our method generalizes across model families (Mistral) and scales to larger MoE architectures (Qwen3-30B-A3B).

\subsection{Efficiency}

\begin{table}[t]
    \centering
    \caption{End-to-end inference latency and peak memory of LLaMA2-7B on a single RTX 3090 GPU at batch size 1: prefill of a 2048-token prompt followed by 128 decoding steps. Both quantized models use per-token asymmetric activation quantization.}
    \label{tab:e2e_latency}
    \resizebox{\linewidth}{!}{
        \begin{tabular}{l cc cc cc cc}
            \toprule
            \multirow{2}{*}{\textbf{Method}} & \multicolumn{2}{c}{\textbf{Prefill}} & \multicolumn{2}{c}{\textbf{Decode}} & \multicolumn{2}{c}{\textbf{E2E}} & \multicolumn{2}{c}{\textbf{Peak memory}} \\
            \cmidrule(lr){2-3}\cmidrule(lr){4-5}\cmidrule(lr){6-7}\cmidrule(lr){8-9}
            & \textbf{ms} & \textbf{Speedup} & \textbf{ms/token} & \textbf{Speedup} & \textbf{ms} & \textbf{Speedup} & \textbf{GB} & \textbf{Saving} \\
            \midrule
            FP16        & 496.7 & 1.00$\times$ & 24.01 & 1.00$\times$ & 3626.6 & 1.00$\times$ & 13.84 & 1.00$\times$ \\
            QuaRot      & 283.4 & 1.75$\times$ & 14.92 & 1.61$\times$ & 2194.0 & 1.65$\times$ & 4.33  & 3.19$\times$ \\
            QuaRot+\alg & 287.3 & 1.73$\times$ & 14.97 & 1.60$\times$ & 2203.5 & 1.65$\times$ & 4.33  & 3.19$\times$ \\
            \midrule
            Overhead of \alg & +1.37\% & & +0.34\% & & +0.43\% & & +0.02\% & \\
            \bottomrule
        \end{tabular}
    }
\end{table}

\paragraph{End-to-end inference}
We measure the end-to-end inference latency of the full LLaMA2-7B model on a single NVIDIA RTX 3090 GPU with our deployment kernels, prefilling a 2048-token prompt and then decoding 128 tokens at batch size 1 (\autoref{tab:e2e_latency}).
Recent rotation-based methods such as SpinQuant~\cite{liu2025spinquant} and DartQuant~\cite{shao2025dartquant} adopt per-token asymmetric activation quantization, and SpinQuant+\alg and DartQuant+\alg in \autoref{tab:main_acc} use the same asymmetric quantizer as their baselines.
We therefore measure the cost of \alg against QuaRot with the same asymmetric quantizer.
\alg adds only 1.37\% to the prefill latency, 0.34\% to the per-token decoding latency, and 0.43\% to the end-to-end latency, and increases the peak memory by only 0.02\%.
Over FP16, QuaRot+\alg achieves 1.73$\times$, 1.60$\times$, and 1.65$\times$ speedups in prefill, decoding, and end-to-end latency, respectively, with a 3.19$\times$ memory saving.

These results confirm that adding \alg does not introduce significant computational overhead, allowing us to maintain the substantial speedup and memory savings benefits of quantization while achieving improved accuracy on language modeling and reasoning tasks. Results with symmetric activation quantization and the overhead of \alg on the down-projection layer alone are provided in \autoref{sec:performance_analysis}.

\subsection{Ablation studies}
\paragraph{Symmetric activation quantization}

To isolate the contribution of asymmetric activation quantization to our method, we compare symmetric and asymmetric per-token quantization for activations on LLaMA2-7B and LLaMA3-8B with GPTQ, averaged over ten seeds, with results shown in~\autoref{tab:ablation_symmetric}.
Under symmetric activation quantization, the effect of \alg varies across models.
Switching to asymmetric per-token activation quantization yields an additional gain for QuaRot+\alg on both models: average zero-shot accuracy improves by 0.99 points on LLaMA3-8B (66.62 $\to$ 67.61) and by 0.32 points on LLaMA2-7B (66.33 $\to$ 66.65), while WikiText2 perplexity drops from 8.09 to 7.88 and from 6.12 to 6.05, respectively.
On LLaMA3-70B (single run, \autoref{tab:ablation_symmetric_70b}), switching QuaRot from symmetric to asymmetric activation quantization changes WikiText2 perplexity, C4 perplexity, and average zero-shot accuracy from 6.60, 12.87, and 70.45 to 6.16, 11.96, and 72.28.
Adding \alg under the same asymmetric quantization gives 5.88, 11.24, and 72.73.

These results confirm that asymmetric activation quantization captures the inherent asymmetry in post-rotation activation distributions and is a key driver of the overall performance improvements reported in our main results.

\begin{table*}[t]
  \newcommand{\hlt}{\rowcolor{tablecolor!50}}
  \centering
  \caption{
    Ablation study on the impact of asymmetric activation quantization with GPTQ. LLaMA2-7B and LLaMA3-8B values are averaged over ten seeds; LLaMA3-70B is a single run, and its symmetric QuaRot row is taken from \autoref{tab:main_acc}.
  }
  \label{tab:ablation_symmetric}
  \label{tab:ablation_symmetric_70b}
  \resizebox{1\linewidth}{!}{
  \begin{tabular}{c l c cc ccccccc}
  \toprule
  \multirow{2}{*}{\textbf{Model}} & \multirow{2}{*}{\textbf{Method}} & \multirow{2}{*}{\textbf{Act asym}} & \multicolumn{2}{c}{\textbf{PPL} ($\downarrow$)} & \multicolumn{7}{c}{\textbf{Acc} ($\uparrow$)} \\
    \cmidrule(lr){4-5} \cmidrule(lr){6-12}
    & & & \textbf{Wiki2} & \textbf{C4} & \textbf{ARC-C} & \textbf{ARC-E} & \textbf{HellaSwag} & \textbf{LAMBADA} & \textbf{PIQA} & \textbf{Winogrande} & \textbf{Avg.} \\
    \midrule
    \multirow{4}{*}{\textbf{2-7B}}
    & QuaRot      & \xmark & 6.12 & 8.31 & 42.41 & 70.26 & 72.62 & 70.61 & 76.99 & 66.35 & 66.54 \\
    & QuaRot+\alg & \xmark & 6.12 & 8.32 & 41.73 & 69.48 & 72.37 & 70.50 & 77.15 & 66.74 & 66.33 \\
    & QuaRot      & \cmark & 6.06 & 8.23 & 42.42 & 70.16 & 72.70 & 70.93 & 77.34 & 66.13 & 66.62 \\
    & QuaRot+\alg & \cmark & 6.05 & 8.23 & 42.08 & 69.89 & 72.70 & 71.13 & 77.20 & 66.91 & 66.65 \\
    \midrule
    \multirow{4}{*}{\textbf{3-8B}}
    & QuaRot      & \xmark & 8.15 & 13.28 & 44.71 & 70.62 & 73.66 & 68.18 & 75.56 & 67.29 & 66.67 \\
    & QuaRot+\alg & \xmark & 8.09 & 13.15 & 45.12 & 70.67 & 73.74 & 67.47 & 75.71 & 67.02 & 66.62 \\
    & QuaRot      & \cmark & 7.94 & 12.92 & 44.99 & 71.30 & 74.00 & 69.08 & 76.03 & 67.55 & 67.16 \\
    & QuaRot+\alg & \cmark & 7.88 & 12.81 & 45.73 & 72.13 & 74.25 & 68.62 & 76.75 & 68.18 & 67.61 \\
    \midrule
    \multirow{3}{*}{\textbf{3-70B}}
    & QuaRot      & \xmark & 6.60 & 12.87 & 49.49 & 74.37 & 77.22 & 71.69 & 78.89 & 71.03 & 70.45 \\
    & QuaRot      & \cmark & 6.16 & 11.96 & 50.94 & 76.47 & 78.66 & 72.04 & 81.07 & 74.51 & 72.28 \\
    & QuaRot+\alg & \cmark & 5.88 & 11.24 & 52.65 & 77.36 & 79.48 & 72.09 & 80.96 & 73.88 & 72.73 \\
    \bottomrule
  \end{tabular}}
\end{table*}

\paragraph{Insensitivity to calibration data}
\begin{wraptable}{r}{0.5\linewidth}
  \vskip -1.2em
  \centering
  \caption{
    Performance of QuaRot+\alg on LLaMA3-8B with GPTQ when the sign adjustment vector is computed from different calibration datasets and seeds.
  }
  \label{tab:calibration_data_insensitivity}
  \resizebox{0.95\linewidth}{!}{
  \begin{tabular}{c c c c}
    \toprule
    \textbf{Calibration Dataset} & \textbf{Wiki2 PPL} & \textbf{C4 PPL} & \textbf{Avg. Acc} \\
    \midrule
    Wiki2 (Seed 0) & 7.85 & 12.74 & 66.66 \\
    Wiki2 (Seed 1) & 7.87 & 12.80 & 67.04 \\
    Wiki2 (Seed 2) & 7.84 & 12.77 & 67.41 \\
    C4 (Seed 0) & 7.88 & 12.78 & 67.26 \\
    C4 (Seed 1) & 7.82 & 12.77 & 66.52 \\
    C4 (Seed 2) & 7.87 & 12.82 & 67.25 \\
    \bottomrule
  \end{tabular}}
\end{wraptable} 
To evaluate the sensitivity of our method to the choice of calibration data, we compare the performance of QuaRot+\alg when using different datasets to compute the sign adjustment vector $\mathbf{n}$. Specifically, we use WikiText-2 and C4 as calibration datasets, each with three different random seeds to select calibration samples. 
As shown in~\autoref{tab:calibration_data_insensitivity}, across the two datasets and three seeds, the WikiText2 and C4 perplexities vary by at most 0.06 and 0.08, and the average accuracy by 0.9 points, indicating that \alg is insensitive to the choice of calibration data.

\paragraph{Adjustment on other linear layers}
\label{ablation:adjustment}
We evaluated sign adjustment at different projection inputs of LLaMA3-8B under GPTQ with per-token asymmetric activations, averaged over six seeds (\autoref{tab:ablation_adjust}). We adjusted the inputs of the q/k/v, out, up/gate, and down projections individually. We also tested combinations within the self-attention block ($\tilde{\mathbf{X}}_{qkv} + \tilde{\mathbf{X}}_{o}$) and the MLP block ($\tilde{\mathbf{X}}_{ug} + \tilde{\mathbf{X}}_{d}$), the inputs to the out and down projections together ($\tilde{\mathbf{X}}_{o} + \tilde{\mathbf{X}}_{d}$), and all inputs together.
Among single positions, adjusting $\tilde{\mathbf{X}}_{d}$ gave the lowest perplexity (7.898 Wiki2, 12.869 C4) and the highest accuracy (67.74), whereas adjusting only $\tilde{\mathbf{X}}_{qkv}$, $\tilde{\mathbf{X}}_{ug}$, or $\tilde{\mathbf{X}}_{o}$ gave 7.957--7.974 Wiki2 PPL and 66.94--67.21 accuracy. Every configuration that included $\tilde{\mathbf{X}}_{d}$ reached a similar perplexity (7.876--7.898), and adjusting all inputs gave the lowest perplexity but no additional accuracy (67.49 vs.\ 67.74).
We therefore apply sign adjustment only at $\tilde{\mathbf{X}}_{d}$, which gave the highest mean accuracy among the evaluated configurations. Some combinations gave lower perplexity, but none gave higher mean accuracy. The down-projection input also requires an online rotation: replacing the Hadamard transform with a learned dense matrix cannot use the FWHT and would add a full matrix multiplication during inference.
\begin{table}[t]
  \newcommand{\hlt}{\rowcolor{tablecolor!50}}
  \centering
  \caption{
  Sign adjustment applied at different projection inputs of LLaMA3-8B with GPTQ weight quantization and per-token asymmetric activation quantization.
  $\tilde{\mathbf{X}}_{qkv}$, $\tilde{\mathbf{X}}_{o}$, $\tilde{\mathbf{X}}_{ug}$, and $\tilde{\mathbf{X}}_{d}$ denote the rotated inputs of the q/k/v, out, up/gate, and down projections.
  Values are mean $\pm$ std over six seeds.
  }
  \label{tab:ablation_adjust}
  \small
  \begin{tabular}{l cc c}
    \toprule
    \multirow{2}{*}{\textbf{Adjusted input}} & \multicolumn{2}{c}{\textbf{PPL} ($\downarrow$)} & \textbf{Avg. Acc} ($\uparrow$) \\
    \cmidrule(lr){2-3} \cmidrule(lr){4-4}
    & \textbf{Wiki2} & \textbf{C4} & \textbf{0-shot} \\
    \midrule
    $\tilde{\mathbf{X}}_{qkv}$ & 7.957 $\pm$ 0.039 & 12.955 $\pm$ 0.060 & 67.21 $\pm$ 0.39 \\
    $\tilde{\mathbf{X}}_{ug}$  & 7.972 $\pm$ 0.050 & 12.964 $\pm$ 0.073 & 67.10 $\pm$ 0.59 \\
    $\tilde{\mathbf{X}}_{o}$   & 7.974 $\pm$ 0.035 & 12.988 $\pm$ 0.052 & 66.94 $\pm$ 0.69 \\
    \hlt $\tilde{\mathbf{X}}_{d}$ & 7.898 $\pm$ 0.019 & 12.869 $\pm$ 0.029 & \textbf{67.74} $\pm$ 0.63 \\
    $\tilde{\mathbf{X}}_{qkv}$ + $\tilde{\mathbf{X}}_{o}$ & 7.941 $\pm$ 0.034 & 12.933 $\pm$ 0.035 & 67.46 $\pm$ 0.38 \\
    $\tilde{\mathbf{X}}_{ug}$ + $\tilde{\mathbf{X}}_{d}$  & 7.885 $\pm$ 0.026 & 12.830 $\pm$ 0.042 & 67.28 $\pm$ 0.43 \\
    $\tilde{\mathbf{X}}_{o}$ + $\tilde{\mathbf{X}}_{d}$   & 7.894 $\pm$ 0.032 & 12.867 $\pm$ 0.042 & 67.61 $\pm$ 0.46 \\
    All $\tilde{\mathbf{X}}$ & \textbf{7.876} $\pm$ 0.021 & \textbf{12.809} $\pm$ 0.049 & 67.49 $\pm$ 0.21 \\
    \bottomrule
  \end{tabular}
\end{table}

\section{Conclusions}
We demonstrate that the Hadamard transform, widely used in existing studies to manage activation outliers through linear transformations, can create fragmented distributions when a massive outlier dominates a token's activation. Although such tokens are rare, the fragmentation they cause substantially increases the quantization error of the layer in which they first appear. To address this fragmentation, we propose a massive outlier-aware sign-adjusted rotation method, named \alg, a computationally efficient method optimized for online transformation. \alg identifies massive outlier channels and calculates sign adjustment vectors using a single forward pass. Our approach quantizes the LLaMA2-7B model in 3 minutes with RTN and 15 minutes with GPTQ. During inference, \alg adds only element-wise sign flips per token per decoder layer. On LLaMA3-70B, QuaRot+\alg achieves a WikiText2 perplexity of 5.88 and an average zero-shot reasoning accuracy of 72.73, surpassing SpinQuant. Our findings suggest that a simple, massive outlier-aware sign adjustment can mitigate this fragmentation of online Hadamard rotation without iterative optimization. Extending \alg to tokens with several massive outlier channels and to layers whose dominant massive outlier channel varies across tokens (\eg the last layers in \autoref{sec:ma_statistics}), for instance by assigning each channel group its own Hadamard transform and sign adjustment vector, is left for future work.
\begin{ack}
This work was supported by the National Research Foundation of Korea (NRF) grant funded by the Korea government (MSIT) (RS-2026-25498006). 
This work was supported by Institute of Information \& communications Technology Planning \& Evaluation (IITP) under the Leading Generative AI Human Resources Development (IITP-2026-RS-2026-25544527) grant funded by the Korea government(MSIT). 
This work was supported by Institute of Information \& communications Technology Planning \& Evaluation (IITP) under the Artificial Intelligence Innovation Human Resources Development (IITP-2026-RS-2026-25547714) grant funded by the Korea government(MSIT).
The authors declare no competing interests.
\end{ack}

\bibliographystyle{plain}
\bibliography{main/custom}

\appendix

\section{Derivation of the bimodality threshold}
\label{sec:bimodality_threshold}
This section details the derivation of the bimodality threshold in~\autoref{eq:bimodality_threshold}.

\paragraph{Lindeberg condition under RHT.}
Fix a token $t$ and an output channel $j$. With independent random signs $r_i \in \{\pm 1\}$, \autoref{eq:decomposition} is a sum of independent zero-mean terms $X_i = x_{t,i} r_i h_{i,j}$, where $|X_i| = |x_{t,i}|/\sqrt{C_{in}}$ is deterministic and the total variance is $s_t^2 = \|\mathbf{x}_t\|_2^2 / C_{in}$. Regarding the sum as a row of a triangular array indexed by $C_{in}$, the Lindeberg condition~\cite{berckmoes2018approximate} requires, for every $\epsilon > 0$,
\begin{equation}
    \frac{1}{s_t^2}\sum_{i} \mathbb{E}\!\left[X_i^2 \, \mathbb{I}\{|X_i| > \epsilon s_t\}\right]
    = \sum_{i \,:\, x_{t,i}^2 > \epsilon^2 \|\mathbf{x}_t\|_2^2} \frac{x_{t,i}^2}{\|\mathbf{x}_t\|_2^2} \;\to\; 0 .
\end{equation}
Because the indicator is deterministic, the Lindeberg condition holds if and only if $\max_i x_{t,i}^2 / \|\mathbf{x}_t\|_2^2 \to 0$. If the maximum ratio vanishes, the index set becomes empty for any fixed $\epsilon$. Otherwise, choosing $\epsilon^2$ below the limit superior of the maximum ratio keeps the largest term in the sum. If $M_t^2 / \|\mathbf{x}_t\|_2^2$ remains bounded away from zero along the array, the Lindeberg condition fails for the full sum. Since each summand has variance $x_{t,i}^2/C_{in}$, Feller's negligibility condition fails together with it, so the Lindeberg--Feller converse does not apply and this failure alone does not rule out a Gaussian limit. Non-Gaussianity follows instead from the decomposition: along any subsequence with $M_t^2/\|\mathbf{x}_t\|_2^2 \to \delta > 0$, the normalized outlier term is an independent symmetric two-point variable with variance $\delta$, and by Cram\'er's decomposition theorem~\cite{cramer1936eigenschaft} a sum of independent variables with a non-Gaussian component cannot be Gaussian. A finite magnitude-based massive outlier label alone does not establish this asymptotic property.

Excluding the outlier channel, the residual sum has variance $\sigma_{rest}^2 = \frac{1}{C_{in}}\sum_{i \neq c_{max}} x_{t,i}^2$ and admits a Gaussian approximation when $\max_{i\ne c_{max}}x_{t,i}^2/\sum_{i\ne c_{max}}x_{t,i}^2\to0$. The outlier term $M_t r_{c_{max}} h_{c_{max},j}$ takes the values $\pm M_t/\sqrt{C_{in}}$ with probability $1/2$ each, independently of the residual sum, which gives the approximate equally weighted mixture in~\autoref{eq:bimodal_mixture}.

For a fixed draw of $\mathbf{r}$, a balanced $c_{max}$-th row of $\mathbf H$ assigns the outlier contribution $+M_t/\sqrt{C_{in}}$ to half of the output channels and $-M_t/\sqrt{C_{in}}$ to the other half. Approximating the residuals within each group as Gaussian gives the two-component description of the channel histogram. \alg aligns the signs of this row, converting these two shifts into a single constant contribution $M_t r_{c_{max}}/\sqrt{C_{in}}$ across all output channels. This is how sign adjustment removes the selected outlier's sign-induced split.

\paragraph{Bimodality of the mixture.}
Under the two-Gaussian approximation, let $\mu = |M_t|/\sqrt{C_{in}}$, $\sigma = \sigma_{rest}>0$, and let $\varphi_\sigma$ denote the density of $\mathcal{N}(0,\sigma^2)$. The density $f(y) = \frac{1}{2}\varphi_\sigma(y-\mu) + \frac{1}{2}\varphi_\sigma(y+\mu)$ is symmetric about zero, so $f'(0)=0$. Using $\varphi_\sigma''(y) = (y^2/\sigma^4 - 1/\sigma^2)\,\varphi_\sigma(y)$,
\begin{equation}
    f''(0) = \frac{1}{\sigma^2}\left(\frac{\mu^2}{\sigma^2} - 1\right)\varphi_\sigma(\mu).
\end{equation}
Hence $y=0$ is a local minimum between two modes if and only if $\mu > \sigma$, and the mixture is unimodal for $\mu \le \sigma$~\cite{schilling2002bimodal}. Substituting $\mu$ and $\sigma$ gives
\begin{equation}
    \frac{|M_t|}{\sqrt{C_{in}}} > \sigma_{rest}
    \iff M_t^2 > \sum_{i \neq c_{max}} x_{t,i}^2
    \iff \rho_t := \frac{M_t^2}{\sum_{i \neq c_{max}} x_{t,i}^2} > 1,
\end{equation}
where $\rho_t$ denotes the dominance ratio of the outlier channel. Since $\|\mathbf{x}_t\|_2^2 = M_t^2 + \sum_{i \neq c_{max}} x_{t,i}^2$, the condition $\rho_t > 1$ is equivalent to $M_t^2 > \|\mathbf{x}_t\|_2^2/2$, which at most one channel per token can satisfy. Below the threshold, the approximating mixture is unimodal. This threshold is not a necessary or sufficient test for bimodality of an arbitrary empirical activation histogram. In particular, uniqueness of an energy-dominant channel does not exclude multiple sources of fragmentation.

\paragraph{Numerical example.}
We relate the threshold to the massive outlier token in~\autoref{tab:activation_distribution_statistics}, taken from the down-projection input of the 2nd decoder layer of LLaMA2-7B ($C_{in} = 11{,}008$), where $M_t = 1510$. After \alg, the selected Hadamard row is constant and all other rows are orthogonal to it. Thus, in exact arithmetic, the mean across output channels is $M_t/\sqrt{C_{in}}$ and the population variance across channels is $C_{in}^{-1}\sum_{i\ne c_{max}}x_{t,i}^2=\sigma_{rest}^2$, using the sign convention of the illustrated token. These identities follow from orthogonality and hold for any input, so the agreement of the reported mean of 14.3906 with $1510/\sqrt{11{,}008}\approx14.39$ is a consistency check rather than a test of the mixture model. Using the reported variance of 0.3059 gives $\sigma_{rest}\approx0.553$ and a formal threshold of $\sqrt{C_{in}}\,\sigma_{rest}\approx58$. The observed outlier is about $26\times$ larger ($\rho_t\approx677$), accounting for approximately 99.85\% of the token's energy. These numerical estimates use the rounded table entries.

This token does not satisfy the residual condition of the two-Gaussian model. The original minimum of $-46.16$ implies that a single residual channel carries at least $46.16^2/(C_{in}\cdot0.3059)\approx63\%$ of the residual energy, and the output after \alg spans only about $\pm2.3\,\sigma_{rest}$ around its mean, much narrower than expected for $C_{in}$ Gaussian samples. The separation in~\autoref{fig:fragmented_distribution} does not rely on this approximation. Since the right factor $\operatorname{diag}(\mathbf{n})$ in~\autoref{eq:full_transform_matrix} only flips the signs of output channels of the same transform, each RHT output $|y_{t,j}|$ equals the corresponding output after \alg, which lies in $[13.1328,\,15.6328]$ for every $j$. Hence no channel falls in $(-13.13,\,13.13)$, and the two clusters are separated by a gap of about 26 without any distributional assumption.

\begin{table}[h]
  \centering
  \caption{
    Comparison of activation distribution statistics for a token with a massive outlier and a token without an outlier.
  }
  \label{tab:activation_distribution_statistics}
  \resizebox{0.95\linewidth}{!}{
  \begin{tabular}{c cccc cccc}
    \toprule
    & \multicolumn{4}{c}{\textbf{Massive Outlier Token}} & \multicolumn{4}{c}{\textbf{Non-outlier Token}} \\
    \cmidrule(r){2-5} \cmidrule(r){6-9}
    & Mean & Variance & Min & Max & Mean & Variance & Min & Max \\
    \midrule
    Original & 0.1377 & 207.3750 & -46.1562 & 1510.0000 & -0.0000 & 0.0001 & -0.2783 & 0.5435 \\
    After RHT & 0.0001 & 207.3750 & -15.6328 & 15.6250 & 0.0000 & 0.0001 & -0.0419 & 0.0378 \\
    With \alg & 14.3906 & 0.3059 & 13.1328 & 15.6328 & 0.0000 & 0.0001 & -0.0419 & 0.0370 \\
    \bottomrule
  \end{tabular}}
\end{table}

\section{Effect of sign adjustment on activation distribution}
\label{sec:activation_distribution_statistics}
To empirically validate our theoretical analysis of how sign adjustment affects the activation distribution, we analyze the statistics of activations before and after applying \alg on both a token without massive outliers and a token with a massive outlier from the 2nd decoder layer down projection layer on LLaMA2-7b.
As shown in~\autoref{tab:activation_distribution_statistics}, for the massive outlier token, the standard Randomized Hadamard Transformation (RHT) reduces the min/max range to [-15.6328, 15.6250], but the variance remains extremely high at 207.3750, indicating that the distribution is still highly dispersed. In contrast, applying \alg effectively collapses the distribution into a tight cluster with a mean of 14.3906 and a significantly reduced variance of 0.3059, demonstrating that the sign adjustment successfully mitigates the impact of the massive outlier. For the non-outlier token, applying \alg does not significantly alter the mean, variance, or min/max values compared to RHT, confirming that our method preserves the well-distributed nature of activations when no massive outliers are present.

\begin{table}[h]
  \centering
  \caption{
    Massive outlier tokens at the down projection input. Tokens/seq is the number of massive outlier tokens per 2048-token sequence; Channels/token is the mean number of massive outlier channels per massive outlier token (for ours, the recursive depth). $M_2/M_1$ (median) and the energy share of $M_1$, $M_1^2/\|\mathbf{x}_t\|_2^2$ (median), are over the tokens selected by our criterion. Layers not listed contain no massive outlier token under either criterion.
  }
  \label{tab:ma_statistics}
  \resizebox{0.95\linewidth}{!}{
  \begin{tabular}{llrrrrrr}
    \toprule
     & & \multicolumn{2}{c}{Tokens/seq} & \multicolumn{2}{c}{Channels/token} & & \\
    \cmidrule(lr){3-4} \cmidrule(lr){5-6}
    Model & Layer & Sun et al.~\cite{sun2024massive} & Ours & Sun et al.~\cite{sun2024massive} & Ours & $M_2/M_1$ & $M_1$ energy (\%) \\
    \midrule
    \multirow{2}{*}{LLaMA2-7B}  & 2  & 2.09 & 0.97 & 1.52 & 2.44 & 0.03 & 99.8 \\
                                & 31 & 2.09 & 0.98 & 2.00 & 4.98 & 0.53 & 78.1 \\
    \midrule
    \multirow{3}{*}{LLaMA2-13B} & 4  & 1.09 & 1.09 & 1.00 & 3.00 & 0.08 & 99.0 \\
                                & 39 & 1.09 & 1.09 & 1.00 & 1.00 & 0.05 & 99.3 \\
                                & 40 & 0.23 & 0.05 & 1.20 & 1.00 & 0.34 & 61.2 \\
    \midrule
    \multirow{2}{*}{LLaMA3-8B}  & 2  & 1.00 & 1.00 & 1.00 & 3.17 & 0.12 & 98.5 \\
                                & 32 & 3.19 & 0.12 & 2.07 & 2.00 & 0.62 & 60.7 \\
    \midrule
    \multirow{3}{*}{LLaMA3-70B} & 4  & 1.04 & 1.04 & 1.00 & 4.02 & 0.34 & 86.9 \\
                                & 79 & 1.04 & 0.51 & 2.67 & 4.14 & 0.65 & 63.1 \\
                                & 80 & 2.13 & 0.09 & 1.32 & 1.00 & 0.32 & 70.0 \\
    \midrule
    \multirow{6}{*}{Mistral-7B-v0.3} & 2  & 2.15 & 2.13 & 1.76 & 2.40 & 0.10 & 98.9 \\
                                & 27 & 0.01 & 0.18 & 1.00 & 1.00 & 0.12 & 87.8 \\
                                & 28 & 0.01 & 0.30 & 1.00 & 1.00 & 0.16 & 78.5 \\
                                & 30 & 0.00 & 2.85 & --   & 1.01 & 0.24 & 62.0 \\
                                & 31 & 2.13 & 2.10 & 2.00 & 2.00 & 0.68 & 62.3 \\
                                & 32 & 0.00 & 0.41 & --   & 1.00 & 0.16 & 54.9 \\
    \bottomrule
  \end{tabular}}
\end{table}

\begin{table}[h]
  \centering
  \caption{
    Massive outlier tokens at the down projection input of LLaMA2-70B and Qwen3-30B-A3B, with the same columns as~\autoref{tab:ma_statistics}. For Qwen3-30B-A3B, counts are token-expert pairs: each token is routed to 8 of the 128 experts of a layer, and the experts of a layer are pooled.
  }
  \label{tab:ma_statistics_large}
  \resizebox{0.95\linewidth}{!}{
  \begin{tabular}{llrrrrrr}
    \toprule
     & & \multicolumn{2}{c}{Tokens/seq} & \multicolumn{2}{c}{Channels/token} & & \\
    \cmidrule(lr){3-4} \cmidrule(lr){5-6}
    Model & Layer & Sun et al.~\cite{sun2024massive} & Ours & Sun et al.~\cite{sun2024massive} & Ours & $M_2/M_1$ & $M_1$ energy (\%) \\
    \midrule
    \multirow{16}{*}{LLaMA2-70B} & 3  & 2.05 & 1.09 & 1.00 & 2.04 & 0.13 & 98.4 \\
                                 & 9  & 1.95 & 0.86 & 3.05 & 3.98 & 0.49 & 79.1 \\
                                 & 13 & 0.95 & 0.99 & 1.00 & 1.00 & 0.01 & 99.9 \\
                                 & 60 & 0.01 & 0.05 & 1.00 & 1.43 & 0.29 & 87.5 \\
                                 & 62 & 0.01 & 0.12 & 1.00 & 1.00 & 0.07 & 86.6 \\
                                 & 64 & 0.01 & 0.04 & 1.00 & 1.00 & 0.05 & 88.7 \\
                                 & 65 & 0.01 & 0.05 & 1.00 & 1.14 & 0.15 & 86.3 \\
                                 & 66 & 0.02 & 0.16 & 1.00 & 1.00 & 0.07 & 87.0 \\
                                 & 72 & 0.02 & 1.30 & 1.00 & 1.01 & 0.08 & 87.3 \\
                                 & 73 & 2.13 & 1.48 & 1.00 & 1.00 & 0.03 & 98.6 \\
                                 & 74 & 1.87 & 2.38 & 1.00 & 1.73 & 0.44 & 73.4 \\
                                 & 76 & 2.51 & 2.05 & 1.00 & 1.00 & 0.23 & 86.3 \\
                                 & 77 & 1.12 & 1.09 & 1.97 & 3.00 & 0.21 & 94.2 \\
                                 & 78 & 5.06 & 3.04 & 1.60 & 1.00 & 0.47 & 65.3 \\
                                 & 79 & 1.32 & 0.39 & 1.35 & 1.34 & 0.26 & 62.6 \\
                                 & 80 & 9.59 & 0.26 & 2.58 & 2.00 & 0.67 & 54.2 \\
    \midrule
    \multirow{10}{*}{Qwen3-30B-A3B} & 2  & 0.94 & 0.88 & 1.00 & 1.04 & 0.01 & 99.9 \\
                                    & 3  & 1.02 & 1.02 & 1.00 & 2.08 & 0.02 & 100.0 \\
                                    & 4  & 1.02 & 1.02 & 1.00 & 1.84 & 0.00 & 100.0 \\
                                    & 42 & 0.02 & 0.04 & 1.00 & 1.00 & 0.12 & 96.0 \\
                                    & 43 & 0.00 & 2.38 & --   & 1.06 & 0.26 & 77.1 \\
                                    & 44 & 0.00 & 9.72 & --   & 1.05 & 0.25 & 73.2 \\
                                    & 45 & 0.00 & 11.83 & --  & 1.02 & 0.26 & 69.7 \\
                                    & 46 & 0.01 & 36.17 & 1.00 & 1.01 & 0.35 & 55.6 \\
                                    & 47 & 0.12 & 8.87 & 1.00 & 1.11 & 0.30 & 63.5 \\
                                    & 48 & 2.83 & 726.31 & 1.06 & 1.43 & 0.44 & 71.3 \\
    \bottomrule
  \end{tabular}}
\end{table}

\begin{figure}[h]
    \centering
    \includegraphics[width=0.95\linewidth]{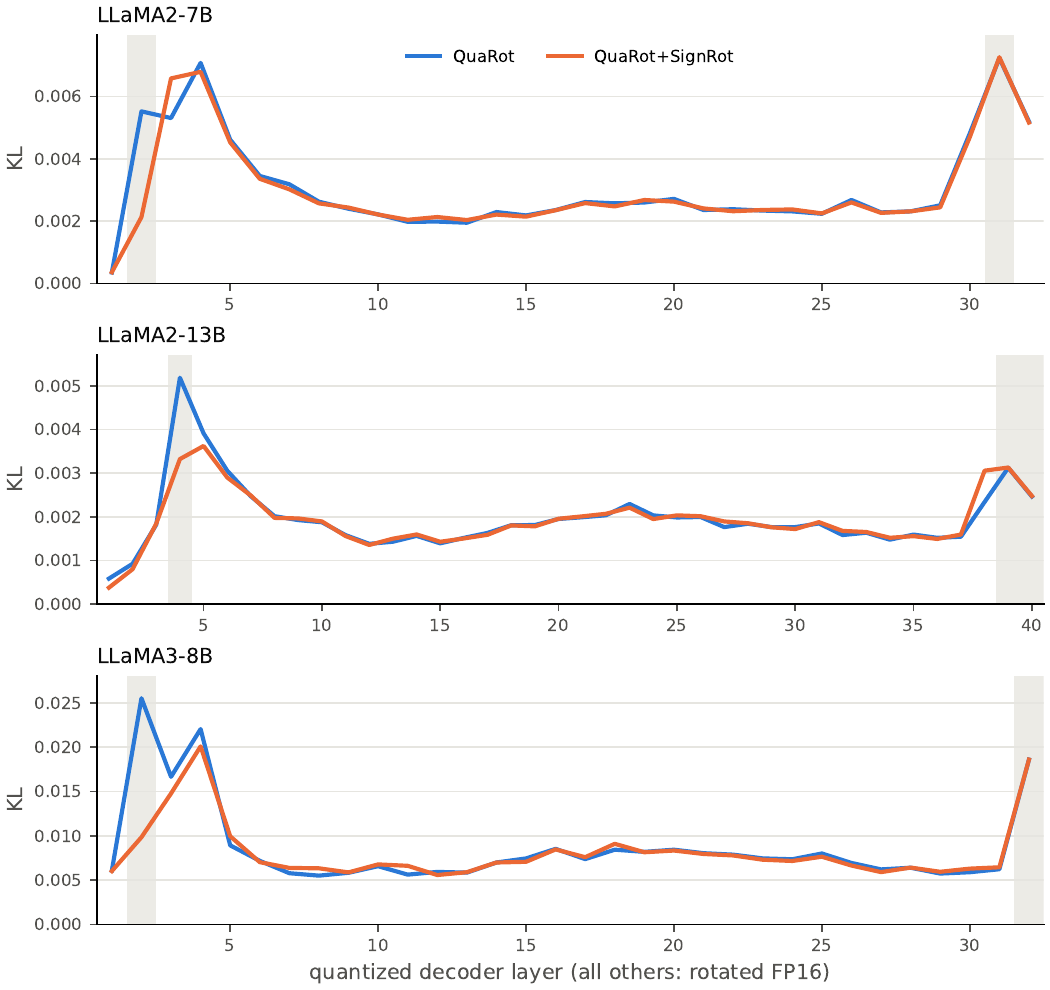}
    \caption{Final KL divergence when only one decoder layer is quantized (GPTQ W4A4, all other layers in rotated FP16). Gray bands mark the layers with massive outlier tokens under our criterion.}
    \label{fig:single_layer_kl}
\end{figure}

\begin{table}[h]
  \centering
  \caption{
    Single-layer quantization KL at the massive outlier layers (GPTQ W4A4). Rank is among all layers of the model with QuaRot; for the 70B models, only 27 layers are measured, and only the first massive outlier layer is listed. $\Delta$ is the change from QuaRot to QuaRot+\alg, over all tokens and over tokens without massive outliers.
  }
  \label{tab:single_layer_kl}
  \resizebox{0.8\linewidth}{!}{
  \begin{tabular}{llrrrrr}
    \toprule
    Model & Layer & Rank & QuaRot & QuaRot+\alg & $\Delta$ all (\%) & $\Delta$ others (\%) \\
    \midrule
    \multirow{2}{*}{LLaMA2-7B}  & 2  & 3 & 0.0055 & 0.0021 & $-61.4$ & $-63.5$ \\
                                & 31 & 1 & 0.0072 & 0.0073 & $+0.4$  & $+4.2$ \\
    \midrule
    \multirow{3}{*}{LLaMA2-13B} & 4  & 1 & 0.0052 & 0.0033 & $-36.0$ & $-31.1$ \\
                                & 39 & 3 & 0.0031 & 0.0031 & $+0.1$  & $+0.3$ \\
                                & 40 & 6 & 0.0025 & 0.0025 & $+0.5$  & $+0.6$ \\
    \midrule
    \multirow{2}{*}{LLaMA3-8B}  & 2  & 1 & 0.0255 & 0.0098 & $-61.4$ & $-62.4$ \\
                                & 32 & 3 & 0.0186 & 0.0186 & $0.0$   & $+1.5$ \\
    \midrule
    LLaMA2-70B                  & 3  & 17 & 0.0017 & 0.0007 & $-56.4$ & $-58.9$ \\
    LLaMA3-70B                  & 4  & 15 & 0.0178 & 0.0069 & $-61.1$ & $-61.8$ \\
    \bottomrule
  \end{tabular}}
\end{table}

\begin{table}[h]
  \centering
  \caption{
    Massive outlier layers within the first sixth of each network (down projection input; the first layer of each model in bold). Tokens/seq is per 2048-token sequence; Top channels is the number of distinct largest channels $c_{max}$ over these tokens (per expert for Qwen3-30B-A3B). $\mu_t > \epsilon_t$ and the range ratio (median of the per-token range before over after \alg) follow~\autoref{sec:multi_ma}, measured on 127 WikiText-2 sequences with the sign adjustment vector calibrated on a separate sequence.
  }
  \label{tab:early_ma_layers}
  \resizebox{0.85\linewidth}{!}{
  \begin{tabular}{lrrrrr}
    \toprule
    Model & Layer & Tokens/seq & Top channels & $\mu_t > \epsilon_t$ (\%) & Range ratio \\
    \midrule
    LLaMA2-7B       & \textbf{2} / 32  & 0.97 & 1 & 100.0 & 11.34$\times$ \\
    LLaMA2-13B      & \textbf{4} / 40  & 1.09 & 1 & 100.0 & 7.00$\times$ \\
    \multirow{3}{*}{LLaMA2-70B} & \textbf{3} / 80 & 1.09 & 1 & 100.0 & 6.05$\times$ \\
                    & 9 / 80  & 0.87 & 1 & 99.1  & 2.26$\times$ \\
                    & 13 / 80 & 0.99 & 1 & 100.0 & 9.40$\times$ \\
    LLaMA3-8B       & \textbf{2} / 32  & 1.00 & 1 & 100.0 & 7.47$\times$ \\
    LLaMA3-70B      & \textbf{4} / 80  & 1.04 & 1 & 100.0 & 2.63$\times$ \\
    Mistral-7B-v0.3      & \textbf{2} / 32  & 2.13 & 1 & 100.0 & 5.53$\times$ \\
    \multirow{3}{*}{Qwen3-30B-A3B} & \textbf{2} / 48 & 0.87 & 1 & 100.0 & 9.51$\times$ \\
                    & 3 / 48  & 1.02 & 1 & 100.0 & 43.60$\times$ \\
                    & 4 / 48  & 1.02 & 1 & 100.0 & 885.36$\times^\dagger$ \\
    \bottomrule
  \end{tabular}}
  \par\smallskip{\footnotesize $^\dagger$Nearly one-hot tokens: after \alg their range is close to zero.\par}
\end{table}

\section{Statistics of massive outliers and their impact on quantization}
\label{sec:ma_statistics}
We run the FP16 model on 128 WikiText-2 sequences of 2048 tokens and record the input of each down projection layer before rotation.
Let $M_1 \geq M_2 \geq \dots$ denote the magnitudes of the largest channels of a token $\mathbf{x}_t \in \mathbb{R}^{C_{in}}$, so that $M_1 = M_t$ in~\autoref{sec:fragmentation}.

\paragraph{Two criteria for a massive outlier token.}
Sun et al.~\cite{sun2024massive} mark a channel as massive when $|x_{t,c}| > 100$ and $|x_{t,c}| > 1000 \times \mathrm{median}_c |x_{t,c}|$.
This criterion identifies large activations, but it is not tied to the condition that causes the bimodal fragmentation in~\autoref{eq:bimodal_mixture}.
After rotation, a token becomes bimodal only when its largest channel outweighs all other channels combined,
\begin{equation}
\rho_1 = \frac{M_1^2}{\sum_{i \neq c_{max}} x_{t,i}^2} > 1,
\label{eq:rho1}
\end{equation}
because the other channels are spread over all $C_{in}$ coordinates and must be compared with $M_1$ as a whole, not with the per-channel median.
We therefore count a token as a massive outlier token when \autoref{eq:rho1} holds and $M_1$ is at least half of the largest magnitude observed in that layer (layer peak $\geq 100$), which excludes low-amplitude sparse tokens.
We apply the same test recursively: a token has depth $k$ when $\rho_j = M_j^2 / \sum_{i > j} x_{t,i}^2 > 1$ for $j = 1, \dots, k$, \ie the $j$-th channel still dominates after the first $j-1$ are removed, and each additional level splits the rotated distribution once more. We refer to these $k$ channels as the massive outlier channels of the token, ordered as the dominant, second, third, \dots\ massive outlier channel.

\paragraph{The magnitude threshold does not track the split.}
As shown in~\autoref{tab:ma_statistics} and~\autoref{tab:ma_statistics_large}, massive outliers are concentrated in the first few and the last few layers: two to six layers with at most about three tokens per sequence in the models up to 13B and in LLaMA3-70B, and more layers in LLaMA2-70B (16) and Qwen3-30B-A3B (10). In the last layer of Qwen3-30B-A3B, 726 token-expert pairs per sequence pass our criterion.

In the first massive outlier layers, the tokens selected by the two criteria share the same largest channel. Elsewhere, the fixed magnitude threshold does not follow the split condition in either direction.
It selects tokens that do not split: in the last layer of LLaMA3-8B it selects 3.19 tokens per sequence, of which only 5.1\% are bimodal after rotation.
It also misses tokens that split but stay below the threshold: in layer~30 of Mistral-7B-v0.3 it selects no token, while 2.85 tokens per sequence pass our criterion.
The miss is more pronounced when $C_{in}$ is small: in the expert down projection of Qwen3-30B-A3B ($C_{in} = 768$), 2.4--36 token-expert pairs per sequence pass our criterion in layers 43--47, of which the magnitude criterion selects at most 0.12.

\paragraph{The first massive outlier layer.}
\autoref{tab:early_ma_layers} lists the massive outlier layers within the first sixth of each network, including LLaMA2-70B and Qwen3-30B-A3B. In every model, the first such layer holds about one massive outlier token per sequence (two in Mistral-7B-v0.3), and all of these tokens share the same largest channel. For all of them, the aligned outlier shift exceeds the largest residual output ($\mu_t > \epsilon_t$, \autoref{sec:multi_ma}), and \alg narrows their per-token range by $2.6$--$11.3\times$ (median).

\paragraph{The largest channel dominates.}
Under our criterion, a massive outlier token in the first massive outlier layer of the LLaMA models carries 2.0--4.0 massive outlier channels on average, but the second massive outlier channel is small.
In the first massive outlier layer of every model except LLaMA3-70B, $M_2/M_1$ is 0.01--0.13 and $M_1$ holds 98.4--99.9\% of the token energy, so aligning the sign of $M_1$ alone, as \alg does, removes almost all of the bimodality.
In LLaMA3-70B, $M_2/M_1$ is 0.34 and $M_1$ holds 87\%; \autoref{sec:multi_ma} shows that \alg still removes the split of $M_1$ in this case.

The exceptions are the last layers of LLaMA2-7B, LLaMA2-70B, LLaMA3-8B, LLaMA3-70B, and Mistral-7B-v0.3 ($M_2/M_1 = 0.53$--$0.68$, $M_1$ holds 54--78\%), where a single sign vector cannot fold both channels.

\paragraph{Massive outlier layers dominate the quantization error.}
To measure how much each layer contributes, we quantize a single decoder layer while keeping all other layers in rotated FP16, and measure the KL divergence of the final output from the FP16 model on WikiText-2 (8 sequences of 2048 tokens, GPTQ W4A4).
The KL of quantizing all layers is close to the sum of these single-layer KLs (ratio 0.96, 1.07, and 0.93 for LLaMA2-7B, LLaMA2-13B, and LLaMA3-8B), so the single-layer KL indicates how much each layer contributes.
\autoref{fig:single_layer_kl} and~\autoref{tab:single_layer_kl} show that the first massive outlier layer is among the three most expensive layers with QuaRot in LLaMA2-7B, LLaMA2-13B, and LLaMA3-8B, at 2.3--3.6$\times$ the median of the layers without massive outliers.

\paragraph{\alg removes the error of the first massive outlier layer.}
As shown in~\autoref{tab:single_layer_kl}, \alg reduces the single-layer KL of the first massive outlier layer by 36--61\%, while the curves of the two methods nearly coincide at layers without massive outliers.
The reduction is not limited to the massive outlier tokens themselves: the KL of all other tokens drops by 31--64\%.
Since \alg barely changes the KL of layers without massive outliers, this suggests that the error on the massive outlier token, which serves as an attention sink, propagates to every other token through attention in the subsequent layers.
At the last layers, \alg leaves the final KL unchanged, as there is no later layer through which the error of the massive outlier token can reach other tokens.
The gain carries over to quantizing all layers. Averaged over random seeds (11 for LLaMA2-7B and LLaMA3-8B, 5 for LLaMA2-13B and Mistral-7B-v0.3), \alg reduces the KL of the fully quantized model by 2.0--4.5\% with GPTQ W4A4 and by 2.9--6.5\% with W16A4 in every model, and enabling \alg only at the first massive outlier layer already yields 1.1--4.8\% and 2.3--6.4\%, respectively.

\section{Sign adjustment with multiple massive outlier channels}
\label{sec:multi_ma}

A massive outlier token can carry more than one massive outlier channel (recursive depth larger than one in~\autoref{sec:ma_statistics}). We show that \alg still removes the split induced by the selected channel without any assumption on the other channels.
Fix a token $t$ and write~\autoref{eq:decomposition} as
\begin{equation}
\label{eq:multi_ma_decomposition}
    y_{t,j} = s_j\,\mu_t + e_{t,j}, \qquad s_j = \operatorname{sign}\!\left(M_t r_{c_{max}} h_{c_{max},j}\right) \in \{\pm 1\}, \qquad \mu_t = \frac{|M_t|}{\sqrt{C_{in}}},
\end{equation}
where the residual $e_{t,j}$ collects every other channel, including the other massive outlier channels, and let $\epsilon_t = \max_j |e_{t,j}|$.

\paragraph{The split of the selected channel is removed for any residual.}
\alg multiplies output $j$ by $n_j = \operatorname{sign}(h_{c_{max},j})$ (\autoref{eq:sign_adjust}), so $n_j s_j = \operatorname{sign}(M_t r_{c_{max}})$ is the same for all $j$ and the adjusted outputs are $\operatorname{sign}(M_t r_{c_{max}})\,\mu_t + n_j e_{t,j}$. All of them lie in an interval of length $2\epsilon_t$ around a common shift. This uses only $|n_j e_{t,j}| \le \epsilon_t$: the residual may have any shape, and the other massive outlier channels may still split it, but only within this interval. Before the adjustment, if $\mu_t > \epsilon_t$, no output falls in $(-(\mu_t-\epsilon_t),\,\mu_t-\epsilon_t)$, and the range is at least $2(\mu_t-\epsilon_t)$ whenever the $c_{max}$-th row contains both signs. Hence the per-token range shrinks by a factor of at least $\mu_t/\epsilon_t - 1$.
The energy-dominance condition in~\autoref{eq:bimodality_threshold} therefore keeps its role when a token carries several massive outlier channels: \alg removes the split of the dominant massive outlier channel, and the remaining channels only enter through $\epsilon_t$. Folding the remaining channels as well, \eg by assigning each channel group its own Hadamard transform and sign adjustment vector, is left for future work.
For example, every massive outlier token in the first massive outlier layer of LLaMA3-70B has more than one massive outlier channel, with $M_2/M_1 \approx 0.34$ (\autoref{tab:ma_statistics}). Yet \alg still reduces their per-token range by $2.6\times$ (median over 127 WikiText-2 sequences, with the sign adjustment vector calibrated on a separate sequence).

\section{Quantization configurations}
\label{sec:detailed_quant_config}
\begin{table}[h]
  \centering
  \caption{
    Quantization configurations of the methods in \autoref{tab:main_acc}. Within each pair, the baseline and +\alg use the same rotation matrices. The clipping ratios of +\alg are listed in \autoref{tab:clip_config}.
  }
  \label{tab:quant_config}
  \resizebox{\linewidth}{!}{
  \begin{tabular}{l l c c c}
    \toprule
    \textbf{Method} & \textbf{Rotation} & \textbf{Act. quantization} & \textbf{Weight granularity} & \textbf{GPTQ damping} \\
    \midrule
    QuaRot            & Randomized Hadamard      & Per-token symmetric  & Per-channel & 0.01 \\
    QuaRot+\alg       & Randomized Hadamard      & Per-token asymmetric & Per-channel & 0.01 \\
    \midrule
    SpinQuant         & Released learned rotation & Per-token asymmetric & Per-channel & 0.01 \\
    SpinQuant+\alg    & Released learned rotation & Per-token asymmetric & Per-channel & 0.01 \\
    \midrule
    DartQuant         & Trained rotation          & Per-token asymmetric & Group 128   & 0.1 \\
    DartQuant+\alg    & Trained rotation          & Per-token asymmetric & Group 128   & 0.1 \\
    \bottomrule
  \end{tabular}}
\end{table}

\begin{table}[h]
  \centering
  \caption{
    Clipping ratios of +\alg in \autoref{tab:main_acc} for the activations, keys, and values.
  }
  \label{tab:clip_config}
  \footnotesize
  \begin{tabular}{l c c c}
    \toprule
    \textbf{Method} & \textbf{Act. clip} & \textbf{K clip} & \textbf{V clip} \\
    \midrule
    QuaRot+\alg       & 0.9$^{a}$ & 0.9 & 0.9 \\
    SpinQuant+\alg    & 0.9       & 0.9 & 0.9 \\
    DartQuant+\alg    & 0.9       & 1.0 & 1.0 \\
    \bottomrule
  \end{tabular}\par
  \vspace{0.5ex}
  {\scriptsize
  $^{a}$1.0 on LLaMA3-70B with GPTQ.\par}
\end{table}

\autoref{tab:quant_config} lists the quantization configuration of each method in \autoref{tab:main_acc}, and \autoref{tab:clip_config} lists the clipping ratios of +\alg. All methods use 4-bit symmetric weight quantization and group-wise asymmetric 4-bit KV cache quantization with a group size of the head dimension. QuaRot follows its official setting~\cite{ashkboos2024quarot}, and SpinQuant uses the rotation matrices released by its authors~\cite{liu2025spinquant}. Within each pair, +\alg uses the same rotation matrices, weight quantization, and GPTQ settings as the baseline; it changes only the activation quantizer to per-token asymmetric quantization and the clipping ratios. QuaRot and QuaRot+\alg on Mistral-7B-v0.3 and Qwen3-30B-A3B (\autoref{tab:ablation_mistral}) use the same configurations as on the LLaMA models, with an activation clipping ratio of 0.9 for QuaRot+\alg.

On LLaMA3-70B with GPTQ, QuaRot+\alg uses an activation clipping ratio of 1.0. This exception comes from QuaRot itself rather than from \alg: with asymmetric activation quantization and an activation clipping ratio of 0.9, the official QuaRot implementation already degrades severely on this model (WikiText2 perplexity 10.05, compared with 6.60 with symmetric activations). We therefore use 1.0 for activations in the asymmetric QuaRot configurations on LLaMA3-70B with GPTQ, including the asymmetric QuaRot baseline in~\autoref{tab:ablation_symmetric_70b}.

\section{Full results of main result table}
\label{sec:full_results}
In Tables~\ref{tab:with_quarot},~\ref{tab:with_spinquant} and~\ref{tab:with_dartquant}, we present the full results of the main result table in the paper, including all evaluated results of LLaMA models. These tables provide a comprehensive comparison of our proposed \alg method against various baselines such as QuaRot~\cite{ashkboos2024quarot}, SpinQuant~\cite{liu2025spinquant}, and DartQuant~\cite{shao2025dartquant}, demonstrating the consistent performance improvements achieved by \alg in managing activation outliers and enhancing quantization efficiency.

\begin{table*}[h]
  \newcommand{\hlt}{\rowcolor{tablecolor!50}}
  \centering
  \caption{
    Performance comparison when applying \alg to QuaRot. `+\alg' indicates that our method has been applied.
  }
  \label{tab:with_quarot}
  \resizebox{0.95\linewidth}{!}{
  \begin{tabular}{c l c cc ccccccc}
    \toprule
    \multicolumn{1}{c}{\textbf{Model}} & \multicolumn{1}{l}{\textbf{Method}} & \multicolumn{1}{c}{\textbf{W Quantizer}} & \textbf{Wiki2} & \textbf{C4} & \textbf{ARC-C} & \textbf{ARC-E}   & \textbf{HellaSwag}   & \textbf{LAMBADA}   & \textbf{PIQA} & \textbf{Winogrande} & \textbf{Avg} \\
    \midrule
    
    \multirow{5}{*}{\textbf{2-7B}}
    & FP16          & -    & 5.47  & 7.26 & 46.16 & 74.54 & 75.98 & 73.92 & 79.05 & 69.06 & 69.79 \\
    \cmidrule{2-12}
    & QuaRot        & RTN  & 8.56  & 11.86 & 36.60 & 61.41 & 65.07 & 48.06 & 72.20 & 63.06 & 57.73 \\
    & QuaRot+\alg   & RTN  & 7.93  & 11.21 & 38.31 & 62.84 & 66.60 & 58.63 & 73.61 & 62.59 & 60.43 \\
    \cmidrule{2-12}
    & QuaRot        & GPTQ & 6.10  & 8.32 & 42.32 & 68.35 & 72.53 & 65.40 & 76.33 & 65.11 & 65.01 \\
    & QuaRot+\alg   & GPTQ & 6.09  & 8.31 & 42.83 & 70.08 & 73.26 & 71.73 & 78.07 & 67.17 & 67.19 \\
    \midrule
    
    \multirow{5}{*}{\textbf{2-13B}}
    & FP16          & -    & 4.88  & 6.73 & 49.15 & 77.44 & 79.39 & 76.73 & 80.47 & 72.14 & 72.55 \\
    \cmidrule{2-12}
    & QuaRot        & RTN  & 6.10  & 8.67 & 42.83 & 69.95 & 73.54 & 65.62 & 77.69 & 67.88 & 66.25 \\
    & QuaRot+\alg   & RTN  & 6.00  & 8.48 & 45.05 & 71.68 & 73.66 & 69.96 & 78.18 & 66.54 & 67.51 \\
    \cmidrule{2-12}
    & QuaRot        & GPTQ & 5.40  & 7.54 & 45.48 & 73.27 & 76.03 & 69.01 & 79.05 & 70.64 & 68.91 \\ 
    & QuaRot+\alg   & GPTQ & 5.32  & 7.43 & 48.29 & 74.49 & 77.20 & 74.40 & 78.78 & 71.03 & 70.70 \\
    \midrule

    \multirow{5}{*}{\textbf{2-70B}}
    & FP16          & -    & 3.32  & 5.71 & 57.17 & 81.02 & 83.81 & 79.60 & 82.70 & 77.98 & 77.05 \\
    \cmidrule{2-12}
    & QuaRot        & RTN  & 4.14  & 6.42 & 52.22 & 76.60 & 79.96 & 74.61 & 81.12 & 76.32 & 73.47 \\
    & QuaRot+\alg   & RTN  & 4.05  & 6.32 & 52.56 & 77.36 & 80.74 & 76.65 & 80.58 & 75.30 & 73.86 \\
    \cmidrule{2-12}
    & QuaRot        & GPTQ & 3.79  & 6.12 & 55.46 & 79.76 & 81.58 & 79.35 & 81.83 & 76.09 & 75.68 \\
    & QuaRot+\alg   & GPTQ & 3.74  & 6.04 & 55.20 & 78.28 & 82.59 & 78.88 & 81.23 & 77.90 & 75.68 \\ 
    \midrule

    \multirow{5}{*}{\textbf{3-8B}}
    & FP16          & -    & 6.14  & 9.45 & 53.50 & 77.57 & 79.12 & 75.51 & 80.74 & 72.93 & 73.23 \\
    \cmidrule{2-12}
    & QuaRot        & RTN  & 10.60 & 17.19 & 38.65 & 66.54 & 68.82 & 57.20 & 71.82 & 65.04 & 61.34 \\  
    & QuaRot+\alg   & RTN  & 9.68 & 15.72 & 42.66 & 67.80 & 71.35 & 64.47 & 74.27 & 64.48 & 64.17 \\
    \cmidrule{2-12}
    & QuaRot	      & GPTQ & 8.16  & 13.38 & 45.73 & 70.83 & 72.97 & 62.70 & 75.35 & 67.17 & 65.79 \\
    & QuaRot+\alg   & GPTQ & 7.85  & 12.74 & 44.88 & 71.72 & 74.16 & 68.83 & 76.06 & 67.96 & 67.27 \\
    \midrule

    \multirow{5}{*}{\textbf{3-70B}}
    & FP16          & -    & 2.86  & 7.17  & 64.25 & 85.94 & 84.93 & 79.37 & 84.44 & 80.74 & 79.95 \\
    \cmidrule{2-12}
    & QuaRot        & RTN  & 55.44 & 79.48 & 22.18 & 34.30 & 32.15 & 13.35 & 57.67 & 52.49 & 35.36 \\
    & QuaRot+\alg   & RTN  & 41.47 & 58.26 & 22.78 & 40.74 & 34.52 & 20.70 & 60.77 & 51.61 & 38.52 \\
    \cmidrule{2-12}
    & QuaRot        & GPTQ & 6.60  & 12.87 & 49.49 & 74.37 & 77.22 & 71.69 & 78.89 & 71.03 & 70.45 \\
    & QuaRot+\alg   & GPTQ & 5.88  & 11.24 & 52.65 & 77.36 & 79.48 & 72.09 & 80.96 & 73.88 & 72.73 \\
    \bottomrule
  \end{tabular}}
\end{table*}

\begin{table}[h]
  \newcommand{\hlt}{\rowcolor{tablecolor!50}}
  \centering
  \caption{
    Performance comparison when applying \alg to SpinQuant. All results are reproduced using SpinQuant author-provided rotation matrices. `+\alg' indicates that our method has been applied.
  }
  \label{tab:with_spinquant}
  \resizebox{0.95\linewidth}{!}{
  \begin{tabular}{c l c cc ccccccc}
    \toprule
    \multicolumn{1}{c}{\textbf{Model}} & \multicolumn{1}{l}{\textbf{Method}} & \multicolumn{1}{c}{\textbf{W Quantizer}} & \textbf{Wiki2} & \textbf{C4} & \textbf{ARC-C} & \textbf{ARC-E}   & \textbf{HellaSwag}   & \textbf{LAMBADA}   & \textbf{PIQA} & \textbf{Winogrande} & \textbf{Avg} \\
    \midrule
    \multirow{2}{*}{\textbf{2-7B}}
    & FP16          & -    & 5.47  & 7.26 & 46.16 & 74.54 & 75.98 & 73.92 & 79.05 & 69.06 & 69.79 \\
    \cmidrule{2-12}
    & SpinQuant      & RTN & 7.71 & 11.14 & 36.77 & 64.27 & 66.53 & 60.31 & 74.43 & 61.72 & 60.67 \\
    & SpinQuant+\alg & RTN & 7.69 & 10.94 & 37.97 & 64.48 & 65.88 & 60.88 & 74.76 & 62.04 & 61.00 \\
    \cmidrule{2-12}
    & SpinQuant      & GPTQ & 6.08 & 8.25 & 41.21 & 68.94 & 72.95 & 71.63 & 77.64 & 65.27 & 66.27 \\
    & SpinQuant+\alg & GPTQ & 6.09 & 8.27 & 42.66 & 70.24 & 72.18 & 72.39 & 77.31 & 67.64 & 67.07 \\
    \midrule
    \multirow{2}{*}{\textbf{2-13B}}
    & FP16          & -    & 4.88  & 6.73 & 49.15 & 77.44 & 79.39 & 76.73 & 80.47 & 72.14 & 72.55 \\
    \cmidrule{2-12}
    & SpinQuant      & RTN & 6.03 & 8.42 & 47.18 & 74.66 & 73.32 & 71.78 & 77.80 & 70.88 & 69.27 \\
    & SpinQuant+\alg & RTN & 5.99 & 8.41 & 47.53 & 74.87 & 73.51 & 71.05 & 78.02 & 69.93 & 69.15 \\
    \cmidrule{2-12}
    & SpinQuant      & GPTQ & 5.39 & 7.46 & 46.33 & 74.58 & 76.09 & 74.40 & 78.45 & 69.77 & 69.94 \\
    & SpinQuant+\alg & GPTQ & 5.37 & 7.45 & 47.53 & 75.38 & 76.42 & 74.69 & 79.43 & 71.59 & 70.84 \\
    \midrule
    \multirow{2}{*}{\textbf{2-70B}}
    & FP16          & -    & 3.32  & 5.71 & 57.17 & 81.02 & 83.81 & 79.60 & 82.70 & 77.98 & 77.05 \\
    \cmidrule{2-12}
    & SpinQuant      & RTN & 4.02 & 6.29 & 55.72 & 78.79 & 81.03 & 77.68 & 82.05 & 75.53 & 75.13 \\
    & SpinQuant+\alg & RTN & 4.00 & 6.29 & 56.06 & 79.92 & 80.36 & 77.86 & 81.39 & 76.16 & 75.29 \\
    \cmidrule{2-12}
    & SpinQuant      & GPTQ & 3.73 & 6.04 & 56.66 & 79.71 & 81.81 & 78.73 & 82.15 & 74.82 & 75.65 \\
    & SpinQuant+\alg & GPTQ & 3.72 & 6.04 & 56.14 & 79.76 & 81.45 & 78.92 & 81.45 & 77.03 & 75.79 \\
    \midrule
    \multirow{2}{*}{\textbf{3-8B}}
    & FP16          & -    & 6.14  & 9.45 & 53.50 & 77.57 & 79.12 & 75.51 & 80.74 & 72.93 & 73.23 \\
    \cmidrule{2-12}
    & SpinQuant      & RTN & 9.57 & 15.42 & 41.38 & 68.48 & 71.07 & 62.74 & 75.14 & 66.77 & 64.26 \\
    & SpinQuant+\alg & RTN & 9.41 & 15.42 & 41.81 & 67.76 & 71.14 & 63.07 & 73.99 & 66.14 & 63.99 \\
    \cmidrule{2-12}
    & SpinQuant      & GPTQ & 7.85 & 12.71 & 47.18 & 71.38 & 74.71 & 69.45 & 76.77 & 68.27 & 67.96 \\
    & SpinQuant+\alg & GPTQ & 7.87 & 12.60 & 47.27 & 73.23 & 74.79 & 72.19 & 77.75 & 70.96 & 69.36 \\
    \midrule
    \multirow{2}{*}{\textbf{3-70B}}
    & FP16          & -    & 2.86  & 7.17  & 64.25 & 85.94 & 84.93 & 79.37 & 84.44 & 80.74 & 79.95 \\
    \cmidrule{2-12}
    & SpinQuant      & RTN & 24.33 & 40.84 & 23.29 & 39.23 & 39.12 & 16.51 & 57.62 & 51.54 & 37.88 \\
    & SpinQuant+\alg & RTN & 23.14 & 37.88 & 25.17 & 47.85 & 45.09 & 21.73 & 62.57 & 53.51 & 42.65 \\
    \cmidrule{2-12}
    & SpinQuant      & GPTQ & 6.38 & 13.98 & 53.24 & 77.86 & 74.16 & 70.48 & 79.11 & 72.06 & 71.15 \\
    & SpinQuant+\alg & GPTQ & 5.45 & 10.54 & 53.75 & 77.53 & 79.56 & 74.64 & 79.82 & 75.85 & 73.52 \\
    \bottomrule
  \end{tabular}}
\end{table}

\begin{table*}[h]
  \newcommand{\hlt}{\rowcolor{tablecolor!50}}
  \centering
  \caption{
    Performance comparison when applying \alg to DartQuant. All results are reproduced using rotation matrices trained with the DartQuant method. '+\alg' indicates that our method has been applied.
  }
  \label{tab:with_dartquant}
  \resizebox{0.95\linewidth}{!}{
  \begin{tabular}{c l cc ccccccc}
    \toprule
    \multicolumn{1}{c}{\textbf{Model}} & \multicolumn{1}{l}{\textbf{Method}} & \textbf{Wiki2} & \textbf{C4} & \textbf{ARC-C} & \textbf{ARC-E}   & \textbf{HellaSwag}   & \textbf{LAMBADA}   & \textbf{PIQA} & \textbf{Winogrande} & \textbf{Avg} \\
    \midrule

    \multirow{2}{*}{\textbf{2-7B}}
    & DartQuant      & 5.97 & 8.07 & 42.58 & 69.74 & 73.49 & 71.36 & 77.69 & 67.01 & 66.98 \\
    & DartQuant+\alg & 5.94 & 8.04 & 42.58 & 70.58 & 73.83 & 71.92 & 77.48 & 67.25 & 67.27 \\
    \midrule

    \multirow{2}{*}{\textbf{2-13B}}
    & DartQuant      & 5.32 & 7.39 & 46.59 & 75.63 & 77.26 & 75.02 & 78.51 & 69.46 & 70.41 \\
    & DartQuant+\alg & 5.31 & 7.37 & 47.35 & 75.34 & 76.87 & 74.64 & 79.33 & 70.01 & 70.59 \\
    \midrule

    \multirow{2}{*}{\textbf{2-70B}}
    & DartQuant      & 3.66 & 6.01 & 54.27 & 78.20 & 83.06 & 79.16 & 82.21 & 76.80 & 75.61 \\
    & DartQuant+\alg & 3.66 & 6.00 & 54.86 & 79.00 & 82.90 & 78.92 & 82.10 & 77.35 & 75.86 \\
    \midrule

    \multirow{2}{*}{\textbf{3-8B}}
    & DartQuant      & 7.49 & 12.16 & 48.21 & 73.82 & 75.40 & 69.55 & 76.88 & 71.74 & 69.27 \\
    & DartQuant+\alg & 7.40 & 12.05 & 48.04 & 71.76 & 76.07 & 71.16 & 76.93 & 70.24 & 69.03 \\
    \midrule

    \multirow{2}{*}{\textbf{3-70B}}
    & DartQuant      & 4.93 & 9.45 & 57.94 & 80.68 & 82.95 & 76.17 & 82.10 & 78.53 & 76.39 \\
    & DartQuant+\alg & 4.86 & 9.26 & 57.76 & 80.26 & 83.02 & 76.17 & 81.56 & 77.35 & 76.02 \\
    \bottomrule
  \end{tabular}}
\end{table*}

\begin{table}
\renewcommand{\arraystretch}{1.1}%
    \newcommand{\hlt}{\rowcolor{tablecolor!50}}
    \centering
    \caption{Total quantization time comparison on a single A100 GPU\@. SpinQuant requires 8$\times$A100 GPUs for its rotation training phase.}
    \label{tab:total_quantization_time}
    \resizebox{0.6\columnwidth}{!}{
        \begin{tabular}{l cccc}
            \toprule
            \textbf{Methods}            & LLaMA2-7B & LLaMA2-13B & LLaMA2-70B \\
            \midrule
            \textbf{RTN, QuaRot+\alg}     & 1+2min    & 2+3min    & 5+8min \\ \cdashline{0-3}
            
            \textbf{GPTQ}               & 14min     & 26min     & 1.8h \\
            \textbf{{GPTQ}, QuaRot}      & 1+14min   & 2+26min   & 0.1h+1.8h \\
            \textbf{GPTQ, QuaRot+\alg}    & 1+14min   & 2+27min   & 0.1h+1.8h \\
            \cdashline{0-3}
            \textbf{{GPTQ}, SpinQuant}   & 1.3h+0.2h  & 2.4h+0.4h  & 12h+1.8h \\
            \bottomrule
        \end{tabular}
    }
\end{table}

\subsection{Total quantization time}
\label{appendix:time}
We report the total quantization time comparison in~\autoref{tab:total_quantization_time}. 
RTN+QuaRot+\alg quantizes the LLaMA2-7B and LLaMA2-70B models in only 3 and 13 minutes, respectively, orders of magnitude faster than SpinQuant~\cite{liu2025spinquant}, which requires hours of multi-GPU training. Even when incorporating GPTQ, \alg matches the quantization time of QuaRot~\cite{ashkboos2024quarot}, since the sign adjustment vector $\boldsymbol{n}$ is computed within the GPTQ calibration pass at negligible additional cost.

\section{Performance analysis of \alg}
\label{sec:performance_analysis}

\begin{table}[h]
    \centering
    \caption{End-to-end inference latency and peak memory on a single RTX 3090 GPU at batch size 1 (2048-token prefill, 128 decoding steps) for symmetric and asymmetric per-token activation quantization.}
    \label{tab:e2e_latency_full}
    \resizebox{\linewidth}{!}{
        \begin{tabular}{c l c c c c c}
            \toprule
            \textbf{Model} & \textbf{Method} & \textbf{Act asym} & \textbf{Prefill (ms)} & \textbf{Decode (ms/token)} & \textbf{E2E (ms)} & \textbf{Peak memory (GB)} \\
            \midrule
            \multirow{5}{*}{\textbf{2-7B}}
            & FP16        & -      & 496.7 & 24.01 & 3626.6 & 13.84 \\
            & QuaRot      & \xmark & 228.3 & 14.13 & 2037.6 & 4.24 \\
            & QuaRot+\alg & \xmark & 232.9 & 14.17 & 2047.3 & 4.24 \\
            & QuaRot      & \cmark & 283.4 & 14.92 & 2194.0 & 4.33 \\
            & QuaRot+\alg & \cmark & 287.3 & 14.97 & 2203.5 & 4.33 \\
            \midrule
            \multirow{5}{*}{\textbf{2-13B}}
            & FP16        & -      & \multicolumn{4}{c}{OOM} \\
            & QuaRot      & \xmark & 635.9 & 21.41 & 3376.5 & 7.72 \\
            & QuaRot+\alg & \xmark & 643.2 & 21.56 & 3405.3 & 7.72 \\
            & QuaRot      & \cmark & 722.5 & 22.46 & 3600.1 & 7.72 \\
            & QuaRot+\alg & \cmark & 728.3 & 22.50 & 3610.4 & 7.73 \\
            \bottomrule
        \end{tabular}
    }
\end{table}

\subsection{End-to-end inference latency}
\autoref{tab:e2e_latency_full} reports the end-to-end measurements behind \autoref{tab:e2e_latency} for both symmetric and asymmetric per-token activation quantization.
We run the full model with our deployment kernels: 4-bit matrix multiplications, online Hadamard transforms (with the sign adjustment for \alg), and a 4-bit paged KV cache decoded with FlashInfer.
Weights are quantized with RTN in every run, since latency does not depend on weight values, and each decoding step is replayed from a captured CUDA graph (over a static KV cache for FP16).
Each configuration is measured in two interleaved rounds after a 90-second thermal soak, with five timed repeats per round, and we report the mean.
With symmetric activation quantization, \alg adds 2.02\% to the prefill latency and 0.48\% to the end-to-end latency over QuaRot on LLaMA2-7B, similar to the asymmetric case. On LLaMA2-13B, \alg adds 0.80\% to the prefill latency and 0.29\% to the end-to-end latency with asymmetric activation quantization (1.15\% and 0.85\% with symmetric); the FP16 model does not fit in the 24\,GB memory of the RTX 3090.

\begin{table}[h]
    \centering
    \caption{Inference speed of a single linear layer for a 2048 sequence length. The matrix sizes correspond to the down-projection layers of LLaMA2-7B, LLaMA2-13B, and LLaMA2-70B. We averaged over 100 runs and report the numbers in milliseconds. Each `+' indicates the addition of the corresponding method on top of the previous column (\ie +QuaRot refers to Int4+QuaRot, +SignRot refers to Int4+QuaRot+SignRot)}
    \label{tab:speedup_single_linear}
    \footnotesize
        \begin{tabular}{l c ccccc}
            \toprule
            \textbf{Layer Size}& \textbf{Batch} & \textbf{FP16} & \textbf{Int4} & \textbf{+QuaRot} & \textbf{+\alg} & \textbf{Speedup} \\
            \midrule
            \multirow{6}{*}{\rotatebox{90}{11008x4096}}
            & 1  & 2.59   & 0.73   & 0.77   & 0.85   & 3.05$\times$ \\
            & 2  & 5.06   & 1.45   & 1.52   & 1.65   & 3.07$\times$ \\
            & 4  & 9.85   & 2.92   & 3.04   & 3.29   & 2.99$\times$ \\
            & 8  & 19.78  & 5.73   & 5.97   & 6.44   & 3.07$\times$ \\
            & 16 & 39.71  & 11.27  & 11.79  & 12.76  & 3.11$\times$ \\
            & 32 & 79.58  & 22.53  & 23.51  & 25.39  & 3.13$\times$ \\
            \midrule
            \multirow{6}{*}{\rotatebox{90}{13824x5120}}
            & 1  & 4.00   & 1.02   & 1.14   & 1.24   & 3.23$\times$ \\
            & 2  & 7.94   & 2.08   & 2.31   & 2.48   & 3.20$\times$ \\
            & 4  & 15.62  & 4.17   & 4.55   & 4.85   & 3.22$\times$ \\
            & 8  & 31.01  & 8.26   & 8.94   & 9.54   & 3.25$\times$ \\
            & 16 & 61.88  & 16.37  & 17.62  & 18.81  & 3.29$\times$ \\
            & 32 & 123.98 & 32.61  & 34.93  & 37.19  & 3.33$\times$ \\
            \midrule
            \multirow{6}{*}{\rotatebox[origin=c]{90}{28672x8192}}
            & 1  & 12.61  & 2.74   & 2.77   & 2.91   & 4.33$\times$ \\
            & 2  & 25.84  & 5.63   & 5.63   & 5.89   & 4.39$\times$ \\
            & 4  & 51.03  & 11.53  & 11.42  & 11.89  & 4.29$\times$ \\
            & 8  & 102.30 & 23.10  & 22.96  & 23.91  & 4.28$\times$ \\
            & 16 & 206.30 & 46.19  & 45.53  & 47.46  & 4.35$\times$ \\
            & 32 & 417.19 & 92.16  & 91.50  & 95.45  & 4.37$\times$ \\
            \bottomrule
        \end{tabular}
\end{table}

\subsection{Single linear layer latency}
To isolate the cost of \alg, we benchmark the down-projection layer alone, where all of its additional operations are concentrated.
\autoref{tab:speedup_single_linear} reports the latency (ms) of Int4+QuaRot with and without \alg.
We benchmark the down-projection layer sizes of LLaMA2-7B (11008$\times$4096), LLaMA2-13B (13824$\times$5120), and LLaMA2-70B (28672$\times$8192) at a sequence length of 2048, averaged over 100 runs.
The overhead decreases with larger models, from a maximum of 10.39\% on LLaMA2-7B (batch size 1) to a minimum of 4.12\% on LLaMA2-70B (batch size 4), since the sign adjustment cost becomes negligible compared to the dominant matrix multiplication.
The overhead also diminishes with larger batch sizes as computation becomes the primary bottleneck, and \alg still achieves a 2.99$\times$--4.39$\times$ speedup over FP16 on this layer.

\section{Limitations}
\label{sec:limitations}
Despite the significant performance improvements of \alg, there are still some limitations. First, it is unknown how to efficiently adjust the online transformation when there are no massive outliers. Second, further research is needed to determine whether \alg is effective when using transformation matrices other than the Hadamard matrix. Third, \alg folds only the channel selected during calibration in each layer. The split induced by a second massive outlier channel remains (\autoref{sec:multi_ma}), and in some later layers massive outliers do not occur at a fixed channel: the largest channel differs from token to token (\eg in the last layers of Mistral-7B-v0.3 and Qwen3-30B-A3B), so tokens whose largest channel is not the selected one are not folded. Lastly, the efficiency of the Hadamard transformation decreases when the intermediate dimension is small, such as in MoE architectures.

\section{Broader impact}
\label{sec:broader_impact}
This paper presents \alg, a computationally efficient method optimized for online transformation. \alg can adjust online transformations to manage massive outliers, which cannot be optimized by learning-based methods. By enabling efficient quantization of large language models, \alg can reduce the computational resources required for training and inference, making advanced AI technologies more accessible and environmentally sustainable. However, if misused for harmful purposes, it could have a negative social impact.

\end{document}